\documentclass[sigconf]{acmart}

\usepackage{graphicx}%
\usepackage{rotating}
\usepackage{amsmath,amsfonts}%
\usepackage{enumitem}
\usepackage{color,xcolor}
\usepackage{subcaption}
\usepackage{tablefootnote}
\usepackage{bm}
\usepackage{listings}
\usepackage{multirow}
\usepackage{array}
\usepackage{amsmath}
\usepackage{amsfonts}
\usepackage{float}
\usepackage[ruled,vlined]{algorithm2e}
\usepackage{algorithmic}
\usepackage{amsmath}
\usepackage{colortbl} % \cellcolor
\usepackage{wrapfig}
\usepackage{pifont}
\usepackage{subfiles}
\usepackage{footmisc}
\usepackage{wasysym}
\usepackage{booktabs}

\definecolor{mygray}{gray}{.85}
\definecolor{mygray1}{gray}{.7}
\definecolor{mygray2}{gray}{.93}
\definecolor{mygray-bg}{gray}{0.9}
\definecolor{mygray-u}{gray}{0.92}
\definecolor{mygreen}{RGB}{0, 176, 80}
\definecolor{myblue}{RGB}{0, 176, 240}

\newcommand{\ie}{\textit{i}.\textit{e}.}
\newcommand{\eg}{\textit{e}.\textit{g}.}
\newcommand{\cf}{\textit{cf.}}

\newcommand{\vs}{\textit{vs}.}

\makeatletter
\newcommand{\thickhline}{%
    \noalign {\ifnum 0=`}\fi \hrule height 1pt
    \futurelet \reserved@a \@xhline
}
\makeatother
\newcommand{\cmark}{\ding{51}}%
\newcommand{\xmark}{\ding{55}}%

\AtBeginDocument{%
  \providecommand\BibTeX{{%
    \normalfont B\kern-0.5em{\scshape i\kern-0.25em b}\kern-0.8em\TeX}}}

\copyrightyear{2025} 
\acmYear{2025} 
\setcopyright{acmlicensed}\acmConference[MM '25]{Proceedings of the 33rd
ACM International Conference on Multimedia}{October 27--31, 2025}{Dublin,
Ireland}
\acmBooktitle{Proceedings of the 33rd ACM International Conference on
Multimedia (MM '25), October 27--31, 2025, Dublin, Ireland}
\acmDOI{10.1145/3746027.3755208}
\acmISBN{979-8-4007-2035-2/2025/10}

\begin{document}

\title{Compositional Zero-shot Learning via Progressive Language-based
Observations}

\author{Lin Li}
\orcid{0000-0002-5678-4487}
\affiliation{%
  \institution{AI Chip Center for Emerging Smart Systems}
  \city{Hong Kong}
  \country{China}}
\affiliation{%
  \institution{
 The Hong Kong University of Science and Technology 
  }
  \city{Hong Kong}
  \country{China}}
\email{lllidy@ust.hk}

\author{Guikun Chen}
\orcid{0000-0002-9227-007X}
\affiliation{%
  \institution{Zhejiang University}
  \department{College of Computer Science and Technology}
  \city{Hangzhou}
  \country{China}}
\email{guikunchen@gmail.com}

\author{Zhen Wang}
\orcid{0009-0008-1091-7994}
\affiliation{%
  \institution{The Hong Kong University of Science and Technology}
  \city{Hong Kong}
  \country{China}}
\email{zwangjr@connect.ust.hk}

\author{Jun Xiao}
\orcid{0000-0002-6142-9914}
\affiliation{%
\institution{Zhejiang University}
\department{College of Computer Science and Technology}
\city{Hangzhou}
\country{China}}
\email{junx@cs.zju.edu.cn}

\author{Long Chen}
\authornote{Long Chen is the corresponding author.}
\orcid{0000-0001-6148-9709}
\affiliation{%
  \institution{The Hong Kong University of Science and Technology}
  \city{Hong Kong}
  \country{China}}
\email{longchen@ust.hk}

\renewcommand{\shortauthors}{Lin Li, Guikun Chen, Zhen Wang, Jun Xiao, \& Long Chen}

\begin{abstract}
Compositional zero-shot learning aims to recognize unseen state-object compositions by leveraging known primitives (state and object) during training. However, effectively modeling interactions between primitives and generalizing knowledge to novel compositions remains a perennial challenge. There are two crucial factors: large \textbf{object-conditioned} and \textbf{state-conditioned} variance, \ie, the appearance of states (or objects) can vary significantly when combined with different objects (or states). 
For instance, the state ``\texttt{old}'' can signify vintage design for a ``\texttt{car}'' or advanced age for a ``\texttt{cat}''.
In this paper, we argue that these variances can be mitigated by predicting composition categories based on \textit{salient} observation cues. Therefore, we propose Progressive Language-based Observations (\textbf{PLO}), which can automatically determine the order of observation cues. These ``observation cues'' comprise a series of primitive concepts or graduated descriptions that allow the model to understand image content in a step-by-step manner. Specifically, PLO adopts pre-trained vision-language models (VLMs) to empower the model with observation capabilities. We further devise two variants: a two-step method (PLO-VLM) with a pre-observing classifier dynamically selecting the order of primitive concept-based cues, and a multi-step approach (PLO-LLM) using large language models (LLMs) to craft graduated description-based cues.
Extensive tests on three datasets show PLO's effectiveness in compositional recognition.\looseness=-1
\end{abstract}




\maketitle

\section{Introduction}
\label{sec1}

What enables us humans to recognize new concepts we have never encountered before? It comes down to our capacity to generalize learned knowledge to unseen domains~\citep{mancini2021open,mancini2022learning,li2023distilled}. For instance, when presented with concepts ``\texttt{green apple}'' and ``\texttt{yellow banana}'', one can recognize and imagine the concept ``\texttt{green banana}'' by combining state ``\texttt{green}'' with object ``\texttt{banana}''. Inspired by this innate cognitive ability of humans, Compositional Zero-Shot Learning (\textbf{CZSL}) emerges to tackle the challenge of recognizing \textit{unseen} state-object compositions (\eg, ``\texttt{green banana}'') by leveraging visible primitives (\ie, state and object) in compositional concepts during training and applying the learned knowledge during inference.\looseness=-1

\begin{figure}[!htpb]
  \centering
  \includegraphics[width=0.90\linewidth]{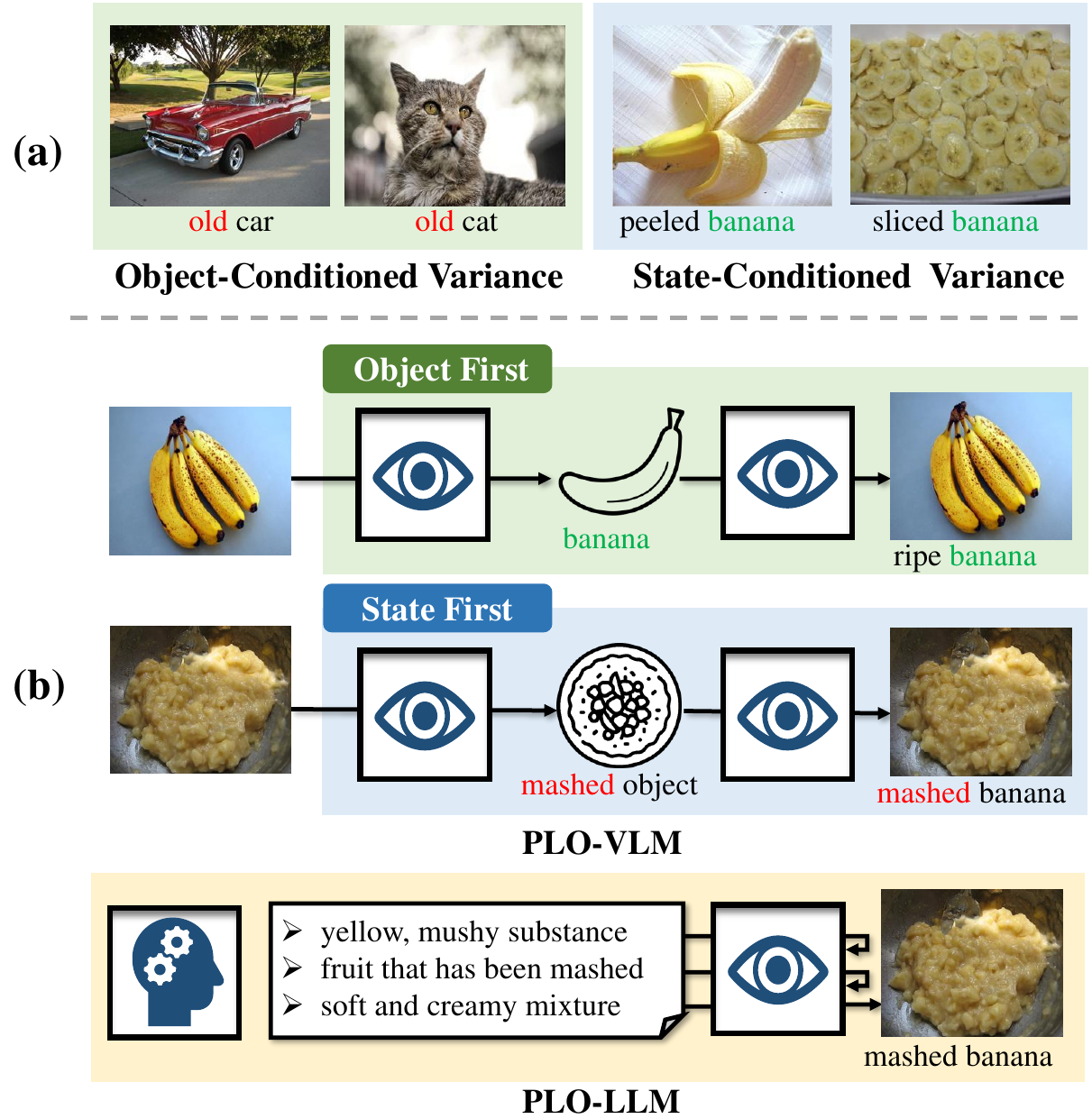}
  \vspace{-1.0em}
  \caption{Illustrations of challenges of CZSL and our PLO. (a) The challenge of object/state-conditioned variation: A perceptible variance emerges in the appearance of state/object primitives when juxtaposed in different compositions. (b) PLO-VLM/LLM: A two/multi-step observation approach dynamically controls the observation order for effective recognition.\looseness=-1} 
  \label{fig:motivation}
  \vspace{-1.0em}
\end{figure}

Effectively modeling the interactions between state and object primitives, as well as inducing the understanding of seen compositions to unseen ones, poses major challenges in CZSL~\citep{mancini2021open,singh2022rethinking,huo2024procc,shao2023counterfactual}. We argue that this task revolves around two key factors: 1) \textbf{Object-conditioned Variance}: Wherein the visual representations of the same state category can vary considerably when different objects are involved. As depicted in Figure~\ref{fig:motivation}(a), consider the state ``\texttt{old}'' in the context of modifying a ``\texttt{car}'' and a ``\texttt{cat}''. For the ``\texttt{car}'', it may refer to a vintage design characterized by classic curves and retro elements, conveying a sense of nostalgia and history. In contrast, for the ``\texttt{cat}'', it denotes the advanced age of the feline, marked by features such as grey fur, which reflect the passage of time and the aging process. 2) \textbf{State-conditioned Variance}: It pertains to the variations in the appearance of an object when combined with different states. In Figure~\ref{fig:motivation}(a), for composition ``\texttt{peeled banana}'', the ``\texttt{banana}'' exhibits a smooth texture and a pale appearance, as the outer peel is removed. In contrast, for ``\texttt{sliced banana}'', the ``\texttt{banana}'' takes on a sliced appearance with visible segments.
Previous CZSL methods often construct separate classifiers for recognizing states and objects simultaneously, overlooking their intrinsic relationship~\citep{nayak2022learning,lu2023decomposed}. Recent efforts have made strides in addressing the \textbf{first factor} by adopting a two-stage method with an object-then-state order~\citep{wang2023learning,huo2024procc,kim2023hierarchical}. Prioritizing the object primitive prediction allows the model to capture salient visual cues (\eg, shapes), thereby enhancing the overall comprehension of compositions. Subsequently, armed with the knowledge of the object primitive, the CZSL model sequentially refines its understanding by classifying the state primitive conditioned on guided object features.\looseness=-1

Nonetheless, we argue that ``\textit{all roads lead to Rome.}'', and the human cognition process will progressively collect different observations for specific compositions, in a \emph{simple to complex} manner~\citep{konkle2024cognitive,shi2024easy,bengio2009curriculum,xiao2020explore,li2023catr}.
In certain cases, such as the composition ``\texttt{ripe banana}'' in Figure~\ref{fig:motivation}(b), the object, ``\texttt{banana}'', possesses salient visual cues that make it easily recognizable due to its curving shape and vibrant yellow color. Once we establish that it is a ``\texttt{banana}'', we can further analyze its state and recognize it as a ``\texttt{ripe banana}'' by observing additional visual cues, \eg, the presence of brown spots on the yellow skin. In contrast, compositions like ``\texttt{mashed banana}'' possess distinct visual features primarily related to the state ``\texttt{mashed}'' rather than the object. The mushy texture becomes the prominent aspect that captures our attention. Consequently, through analysis of extra visual features, \eg, yellow and sticky material, we can refine our recognition and discern it as a ``\texttt{mashed banana}''.\looseness=-1

In this paper, inspired by the human-like cognition process, we propose a novel approach, \underline{\textbf{P}}rogressive \underline{\textbf{L}}anguage-based \underline{\textbf{O}}bservations (\textbf{PLO}) for CZSL. Specifically, PLO automatically determines the order of progressive observation cues in the form of language, building upon the pre-trained vision-language models (VLMs), \eg, CLIP~\citep{radford2021learning}. These ``observation cues'' involve a series of primitive concepts or graduated descriptions that allow the model to observe the image's content step by step. Due to training with image-text pairs, VLMs endow the model with \textit{observing} capability by measuring the similarity between the two modalities within the same space.
For automatic progressive observation, we propose two variants:

\noindent\textbf{PLO-VLM} is a two-step variant that adopts a pre-observing classifier based on VLM to dynamically determine the order of primitive concept-based cues based on image features. With the prior observed primitive knowledge (semantic features from primitive concepts, \eg, ``\texttt{banana}''), we devise a cross-modal attention module that integrates this knowledge for category prediction of the remaining primitive. Since PLO-VLM requires fine-tuning with in-domain visual features, it can be well adapted to specific datasets.

\noindent\textbf{PLO-LLM.} While PLO-VLM advances CZSL, its observing classifier trained on specific datasets might not be generalizable enough for out-of-domain data~\citep{zhu2023debiased,wang2024domain} with distinct images. Thus, we further introduce PLO-LLM, evolving the concept into a more robust multi-step observation scheme. It employs large language models (LLMs) to directly generate domain-agnostic graduated descriptions (\eg, ``yellow, mushy substance'' in Figure~\ref{fig:motivation}(b)) for each composition category, thus obtaining a composition-specific observation order. This method allows us to extract salient features at each observation step according to universal descriptions, boosting the model's ability to more robustly understand and recognize composition categories.

Three prevalent CZSL datasets MIT-States~\citep{isola2015discovering}, UT-Zappos~\citep{naeem2021learning}, and C-GQA~\citep{yu2014fine} are used for evaluation. Extensive results show that our PLO exceeds the current state-of-the-art CZSL methods with significant gains in both closed-world and open-world settings. In addition, cross-domain evaluation is conducted to demonstrate PLO-LLM's excellent generalizability and robustness. 
In summary, the main contributions of our work are four-fold:
\begin{itemize}[leftmargin=*]
\item We argue that the concept of new compositions can vary significantly from the compositions in the dataset, which can be more easily recognized by the salient hints/cues from the concept of easily judged primitive or graduated descriptions from LLMs.
\item We propose an advanced mechanism for CZSL, automatically allocating the observation order in the form of primitive concepts or graduated descriptions, enabling effective prediction of unseen state-object compositions.
\item We introduce two variants: 1) PLO-VLM: using a pre-observing classifier to dynamically determine observation cue order based on image features. 2) PLO-LLM: employing LLMs to directly devise composition-specific graduated descriptions for observation.
\item Extensive results on multiple datasets demonstrate that our PLO outperforms existing CZSL methods, showcasing its effectiveness in recognizing \textit{unseen} state-object compositions.
\end{itemize}

\section{Related Work}

\noindent\textbf{Compositional Zero-Shot Learning (CZSL).}
CZSL is a prominent research area, and state-of-the-art methods can be broadly categorized into two directions: 1) \textbf{Composed CZSL}~\citep{misra2017red,purushwalkam2019task,naeem2021learning,anwaar2022leveraging,mancini2022learning}: It focuses on directly classifying compositional state-object, which projects whole compositional visual features into a shared feature space.
2) \textbf{Decomposed CZSL}~\citep{li2020symmetry,li2022siamese,zhang2022learning,karthik2022kg,khan2023learning,hao2023learning,hu2023leveraging,jiang2023revealing}: It disentangles visual features of simple primitives by designing two classifiers to predict state and object primitives respectively. Notably, recent breakthroughs in CZSL through CLIP-based methods (including both composed and decomposed methods)~\citep{nayak2022learning,lu2023decomposed,lu2023drpt,huang2023troika,zheng2024caila}. These methods leverage frozen CLIP models to acquire textual embeddings of simple primitives, showcasing robust compositionality for zero-shot generalization. Nevertheless, they lack systematic investigation of the intrinsic relation between states and objects, ignoring the priority of visual cues. 
Specifically, Troika~\citep{huang2023troika} adopts a Multi-Path framework to dynamically adjust prompts based on visual features, similar to the idea of the CoCoOP~\citep{zhou2022conditional}. However, it may not fully leverage the inherent relationship between states and objects. In contrast, our PLO innovatively determines the observation cue order through language, allowing for a more nuanced understanding of state-object compositions and advancing the recognition of unseen combinations. This work represents the first effort, as far as we know, that models CZSL in a simple-to-complex manner by progressive language-based observations.

\begin{figure*}[!t]
  \centering
  % \vspace{-3pt}
  \includegraphics[width=0.95\linewidth]{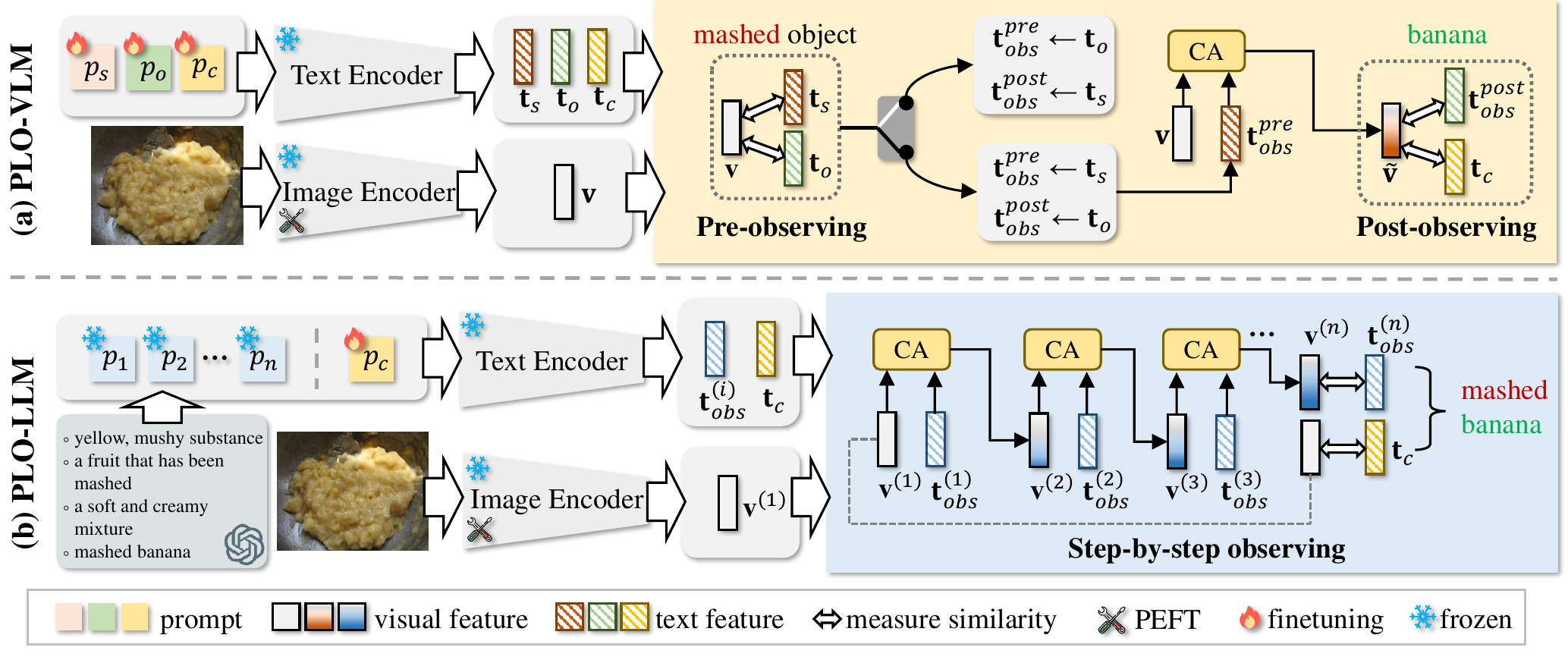}
  \vspace{-1.6em}
  \caption{(a) \textbf{PLO-VLM}: A two-step approach using a pre-observing classifier to dynamically determine the first observation. (b) \textbf{PLO-LLM}: A multi-step approach that observes composition-specific graduated descriptions from LLMs step-by-step.} 
  \label{fig:framework}
  % \vspace{-0.4em}
\end{figure*}

\noindent\textbf{Large Language Models (LLMs) for CV Tasks.}
LLMs have ushered in a new era for natural language understanding and reasoning. There are two main types of solutions to integrate LLMs into CV: 1) Building multi-modal LLMs. Multi-modal pre-training aligns vision and language modalities in various ways~\citep{zang2024contextual,shi2024llmformer,liu2024funnynet,zhu2024vision,ming2024does,wang2025learning}, such as fine-tuning the visual encoder~\citep{tsimpoukelli2021multimodal}, training additional cross-attention layers~\citep{alayrac2022flamingo}, or reducing the size of extra layers and pre-trained lightweight modules~\citep{zhu2023minigpt,li2022blip}. Such \emph{de facto} paradigms require significant computing resources and training data, hindering further applications. 2) Deriving knowledge from LLMs. They query LLMs with meticulously curated prompts to obtain valuable information such as descriptions about category~\citep{li2023zero,menon2023visual,novack2023chils,zhang2023prompt,yang2023language} or synthesized data~\citep{hu2022promptcap}. By such means, vision tasks can enjoy powerful implicit knowledge within LLMs. In this paper, we harness the capabilities of GPT to generate graduated descriptions and facilitate compositional reasoning in CZSL. By leveraging the strengths of LLMs, PLO enhances the understanding of visual compositions and provides a novel approach to address CZSL challenges.

\section{Approach}

\noindent\textbf{Formulation.}
Given an image $I$, CZSL focuses on the recognition of images belonging to composition categories $c \in C$, where the category set $C$ is defined as the Cartesian product of given state categories $S$ = $\{s_1, \ldots, s_{|S|}\}$ and object categories $O$ = $\{o_1, \ldots, o_{|O|}\}$, \ie, $C = S \times O$. During training, the models have access only to a set of seen compositions. \textbf{In closed-world testing}, the model must recognize images from both the seen compositions in $C(s)$ and the unseen compositions in $C(u)$, where $|C(s) \cup C(u)| \ll |C|$. \textbf{In open-world testing}, the model needs to recognize images from any composition within $C$. This scenario presents a more challenging task as the model must handle a broader range of compositions.

In this section, we first introduce how to endow models with \textit{observing} capabilities in Sec.~\ref{sec:3.1}, then describe how to determine the order of step-by-step observation cues for progressive comprehension, \ie, PLO-VLM and PLO-LLM, in Sec.~\ref{sec:3.2} and Sec.~\ref{sec:3.3}.

\subsection{CLIP-based Observing} \label{sec:3.1}

To enable observing ability, PLO builds upon a pre-trained vision-language model: CLIP~\citep{radford2021learning}, which consists of an image encoder $En_v (\cdot)$ and a text encoder $En_t (\cdot)$ capable of mapping image features and semantic features from prompts (natural languages with category information, \eg, ``a photo of ripe banana'') into a shared semantic space. By comparing the similarities between the two modalities in this shared space, we can discern whether language-based observations exist in a given image. 

% \lc{in last paragraph, the $En_t$ refers to the one in CLIP. But in the next paragraph, the $En_t$ is the encoder with the adapter.}

\textbf{1) Image Encoder $En_v(\cdot)$.} We enhance the image encoder of CLIP with some lightweight parameter efficient fine-tuning (PEFT) strategies~\citep{houlsby2019parameter,hu2022lora,zheng2024caila} to effectively handle image features in CZSL. These PEFT strategies
allow the encoder to achieve performance levels comparable to full fine-tuning by transferring knowledge from CLIP while avoiding strong training biases with several learnable parameters. Details of the PEFT strategies are in the Appendix.

Specifically, $En_v (\cdot)$ splits the input image into non-overlapping patches ($N$ in total) along with a pre-trained [CLS] token and positional embeddings, then generates a sequence of patch tokens. After that, the self-attention-based blocks, including the inserted learnable layers, are used to update the token sequence $\mathbf{{\mathcal{V}}}$ = $\{\mathbf{v}_{[\rm{CLS}]}, \mathbf{v}_1, \mathbf{v}_2, \dots, \mathbf{v}_N\}$. By optimizing the parameters of inserted learnable layers during training while keeping the original image encoder frozen, PLO effectively incorporates primitive-specific information to improve observation capabilities. Finally, a linear layer is used to project the output [CLS] token $\hat{\mathbf{v}}_{[\rm{CLS}]}$, yielding the image representation $\mathbf{v}$ in the cross-modal shared space.

\textbf{2) Text Encoder $En_t(\cdot)$.} Following~\citep{lu2023decomposed}, we employ soft prompt tuning, making prompt tokens better adapted to CZSL. Specifically, we formulate the prompt as a set comprising prefix context and category representations. By converting the prompt into learnable embeddings, we provide the model with the flexibility to adapt its language-based observation.
The prompts of state, object, and composition, denoted as $\mathbf{P}_{s}$, $\mathbf{P}_{o}$, and $\mathbf{P}_{c}$, are formulated as:
\begin{equation}
\begin{aligned}
    \mathbf{P}_{s} &= [\mathbf{x}_0; \mathbf{x}_1; \dots; \mathbf{x}_M; \mathbf{x}_{s}; \mathbf{o}], \\
\mathbf{P}_{o} &= [ \mathbf{x}_0; \mathbf{x}_1; \dots; \mathbf{x}_M; \mathbf{x}_{o}], \\
\mathbf{P}_{c}& = [\mathbf{x}_0; \mathbf{x}_1; \dots; \mathbf{x}_M; \mathbf{x}_{s}; \mathbf{x}_{o}], 
\end{aligned}
\end{equation}
where $[\mathbf{x}_0; \dots; \mathbf{x}_M]$ denotes the prefix context with the word embedding of ``a photo of'' as initialization, $\mathbf{x}_s$ and $\mathbf{x}_o$ represent word embedding of the state and object categories, $\mathbf{o}$ is word embedding of the word ``object''. The context length is denoted by $M$. Similar to the token representation in the visual part, these textual prompt embeddings (\ie, context embeddings, category embeddings, and word embedding $\mathbf{o}$) are then converted to learnable embeddings and fed into the text encoder $En_t(\cdot)$ to update the patch tokens $\mathbf{{\mathcal{T}}}$ = $\{\mathbf{t}_{[\rm{CLS}]}, \mathbf{t}_1, \mathbf{t}_2, \dots, \mathbf{t}_M\}$ with self-attention blocks. These learnable embeddings are shared across the three prompts. Subsequently, the output $\hat{\mathbf{t}}_{[\rm{CLS}]}$ of the three types of prompts are fed into a single linear layer to project as text representations $\mathbf{t}_s$, $\mathbf{t}_o$, and $\mathbf{t}_c$. 

These text and image representations are then used in subsequent PLO-VLM and PLO-LLM steps for compositional classification.

\subsection{PLO-VLM} \label{sec:3.2}

As shown in Figure~\ref{fig:framework}(a), PLO-VLM adopts a dynamic two-step observation framework to decide whether to prioritize observing the state or object primitive concept. Next, the prompt features of the selected primitive are integrated into the image using cross-modal attention (CA), which highlights regions of interest related to the observation. By further comparing the similarities between the refined image features and the text features of the remaining primitives, PLO-VLM effectively recognizes state-object compositions.

\noindent\textbf{1) Pre-observing.} The pre-observing classifier recognizes the first primitive by measuring the cosine similarity $\mathcal{S}(\cdot,\cdot)$ between the image representation $\mathbf{v}$ and the text representations of the state primitive $\mathbf{t}_s$ and object primitive $\mathbf{t}_o$:
\begin{equation}
\mathcal{F}^{pre}_{obs}(\mathbf{v}, \mathbf{t}_s, \mathbf{t}_o) = \mathcal{S}(\mathbf{v},\mathbf{t}_s)\oplus\mathcal{S}(\mathbf{v},\mathbf{t}_o),
\end{equation}
where $\oplus$ denotes concatenation. The first observation is the prompt of $argmax(\mathcal{F}^{pre}_{obs}(\cdot))$. Its corresponding text representation is $\mathbf{t}^{pre}_{obs}$. For instance in Figure~\ref{fig:framework}(a), the first observation is the prompt of the state ``\texttt{mashed object}'', and $\mathbf{t}^{pre}_{obs}$ is text representation $\mathbf{t}_{s}$.

\noindent\textbf{2) Post-observing.}
After obtaining the first observation, we use learnable residual CA modules to extract interest features of the first observation for refined image representations $\widetilde{\mathbf{v}}$, which are widely used in methods~\citep{huang2023troika,lu2023decomposed}. The CA module is defined as:
\begin{equation}
    \text{CA}(\mathbf{q},\mathbf{K},\mathbf{V}) = \mathbf{q} + \rm{FFN}(LN(\mathbf{q} + MHA(\mathbf{q},\mathbf{K},\mathbf{V}))),
\end{equation}
where $\mathbf{q}$, $\mathbf{K}$, and $\mathbf{V}$ are the features of query, key, and value. The module FFN, LN, and MHA are feed-forward networks, layer normalization, and multi-head attention. The refined image representations $\widetilde{\mathbf{v}}$ are derived based on the first observation as:
\begin{equation}
\label{eq:CA}
% \begin{aligned}
       \widetilde{\mathcal{V}} = \text{CA}(\mathcal{V}, \mathcal{T}^{pre}_{obs},  \mathcal{T}^{pre}_{obs}), \quad \widetilde{\mathbf{v}} = \text{FC}(\widetilde{\mathcal{V}}),
% \end{aligned}
\end{equation}
where $\mathcal{T}^{pre}_{obs}$ is the output patch tokens of the prompt corresponding to $\mathbf{t}^{pre}_{obs}$. The [CLS] token of $\widetilde{\mathcal{V}}$ is then fed into a linear layer to obtain the refined image representation $\widetilde{\mathbf{v}}$. The refined image representation is utilized to calculate the similarity between the prompt of retain primitive, \eg, object ``\texttt{banana}'' in Figure~\ref{fig:framework}(a), written as: 
\begin{equation}
\begin{aligned}
p(s|I) &= \pi(\mathcal{S}(\widetilde{\mathbf{v}}, \mathbf{t}^{post}_{obs}), \tau), \text{ if  } \mathbf{t}^{post}_{obs} \leftarrow \mathbf{t}_s, \\
p(o|I) &= \pi(\mathcal{S}(\widetilde{\mathbf{v}}, \mathbf{t}^{post}_{obs}), \tau), \text{ if  } \mathbf{t}^{post}_{obs} \leftarrow \mathbf{t}_o. \\
\end{aligned}
\end{equation}
where $\pi(\cdot)$ is the softmax function and $\tau$ denotes the temperature hyper-parameter. The $\mathbf{t}^{post}_{obs}$ represents the text presentation of the prompt of retaining state or object primitive, respectively.
Besides, to ensure that the refined feature encompasses both state and object features, we also compare its similarity with the text presentations of the composition prompt, which is written as:
\begin{equation}
p(c|I) = \pi(\mathcal{S}(\widetilde{\mathbf{v}}, \mathbf{t}_{c}), \tau),
\end{equation}
where the $\mathbf{t}_c$ corresponds to the text representation of the prompt of the composition category.

\subsection{PLO-LLM} \label{sec:3.3}

In Figure~\ref{fig:framework}(b), we employ a multi-step observation process and starts by directly generating a sequence of observation cues (graduated descriptions) for each composition category using an LLM, \ie, GPT~\citep{brown2020language}. Then, the final compositional classification is determined by observing step by step whether cues exist in the image. 

\noindent\textbf{1) Observation Cues Generation.} To elaborate, for each composition category $_{\!}c\in _{\!}C$, a sequence of the hard prompts (observation cues) $\mathcal{P}_c$ = $\{\mathbf{P}_c^{(i)}\}^{n}_{i=1}$ is generated using an LLM with the designed prompt (refer to the Appendix), where $n$ denotes the number of observation steps. Note that these prompts are frozen. For instance, in Figure~\ref{fig:framework}(b), the observation cues are listed in the blue box, \eg, ``yellow, mushy substance'' for ``\texttt{mashed banana}''.

\noindent\textbf{2) Step-by-step Observing.} At each observation step $i$, the image representation $\mathbf{v}^{(i)}$ and the text representation of the current observation cue $\mathbf{t}^{(i)}_{obs}$ are fed into the learnable CA module. The CA module updates the image representation to produce the refined image representation $\widetilde{\mathbf{v}}^{(i)}$:
\begin{equation}
\label{ca_2}
% \begin{aligned}
\setlength\abovedisplayskip{5pt}
\widetilde{\mathcal{V}}^{(i)} \!=\! \text{CA}(\mathcal{V}^{(i)}, \mathcal{T}^{(i)}_{obs}, \mathcal{T}^{(i)}_{obs}), \quad\mathbf{v}^{(i+1)} = \widetilde{\mathbf{v}}^{(i)} = \text{FC}(\widetilde{\mathcal{V}}^{(i)}),
\setlength\belowdisplayskip{5pt}
% \end{aligned}
\end{equation}
where $\widetilde{\mathbf{v}}^{(i)}$ incorporates information from the current observation cue, which is then utilized in the next observation step to calculate the similarity between the image and the next observation cue $\mathbf{t}^{(i+1)}_{obs}$. This process is repeated for all observation steps, iteratively updating the image representation with information from each observation cue. The result of each observation is $\mathcal{S}(\mathbf{v}^{(i)}, \mathbf{t}^{(i)}_{obs})$. 

Since each observation is conditioned on the previous step, the probability of each step is given by:
\begin{equation}
\setlength\abovedisplayskip{5pt}
p(c^{(i)}|I, c^{(1)}, \dots, c^{(i-1)}) = \pi(\mathcal{S}(\mathbf{v}^{(i)}, \mathbf{t}^{(i)}_{obs}), \tau).
\setlength\belowdisplayskip{5pt}
\end{equation}
Furthermore, to facilitate the model in learning semantic representations suitable for CZSL without being solely influenced by hard prompts, we also incorporate the similarity between the original image representation $\mathbf{v}$ and the text representation $\mathbf{t}_c$ of the learnable composition soft prompts. The probability is calculated as:
\begin{equation}
\setlength\abovedisplayskip{5pt}
    p^{Soft}(c|I) = \pi(\mathcal{S}(\mathbf{v}, \mathbf{t}_{c}), \tau).
\setlength\belowdisplayskip{5pt}
\end{equation}
\begin{table*}[!t] \small
  \centering
    \renewcommand\arraystretch{1.0}
    \setlength\tabcolsep{6.0pt}
   \scalebox{1.0}{
    \begin{tabular}{|c||rl|cccc|cccc|cccc|}
    \thickhline
    \multirow{2}[0]{*}{Setting} & \multicolumn{2}{c|}{\multirow{2}[0]{*}{Method}}  & \multicolumn{4}{c|}{MIT-States} & \multicolumn{4}{c|}{UT-Zappos} & \multicolumn{4}{c|}{C-GQA} \\
         
  &   &      & S     & U     & HM    & AUC   & S     & U     & HM    & AUC   & S     & U     & HM    & AUC \\
    \hline
    \hline
    \multirow{13}[0]{*}{\begin{sideways}Closed-World\end{sideways}} & CLIP~\citep{radford2021learning} & $_{\text{ICML'21}}$
    & 30.2  & 46.0  & 26.1  & 11.0  & 15.8  & 49.1  & 15.6  & 5.0   & 7.5   & 25.0  & 8.6   & 1.4  \\
    & CoOp~\citep{zhou2022learning} & $_{\text{IJCV'22}}$  
    & 34.4  & 47.6  & 29.8  & 13.5  & 52.1  & 49.3  & 34.6  & 18.8  & 20.5  & 26.8  & 17.1  & 4.4  \\
    & Co-CGE~\citep{mancini2022learning} & $_{\text{TPAMI'22}}$
    & 46.7 & 45.9 & 33.1 & 17.0 & 63.4 & 71.3 & 49.7 & 36.3 & 34.1 & 21.2 & 18.9 & 5.7 \\
    & ProDA~\citep{lu2022prompt} & $_{\text{CVPR'22}}$
    & 37.4 & 51.7 & 32.7 & 16.1 & 63.7 & 60.7 & 47.6 & 32.7 & - & - & - & - \\
    & CSP~\citep{nayak2022learning}  & $_{\text{ICLR'23}}$
    & 46.6  & 49.9  & 36.3  & 19.4  & 64.2  & 66.2  & 46.6  & 33.0  & 28.8  & 26.8  & 20.5  & 6.2  \\
    & DFSP(i2t)~\citep{lu2023decomposed} & $_{\text{CVPR'23}}$ 
    & 47.4  & 52.4  & 37.2  & 20.7  & 64.2  & 66.4  & 45.1  & 32.1  & 35.6  & 29.3  & 24.3  & 8.7  \\
    & DFSP(BiF)~\citep{lu2023decomposed} & $_{\text{CVPR'23}}$ 
    & 47.1  & 52.8  & 37.7  & 20.8  & 63.3  & 69.2  & 47.1  & 33.5  & 36.5  & 32.0  & 26.2  & 9.9  \\
    & DFSP(t2i)~\citep{lu2023decomposed} & $_{\text{CVPR'23}}$ 
    & 46.9  & 52.0  & 37.3  & 20.6  & 66.7  & 71.7  & 47.2  & 36.0  & 38.2  & 32.0  & 27.1  & 10.5  \\
    & GIPCOL~\citep{xu2023gipcol} & $_{\text{WACV'24}}$
    & 48.5 & 49.6 & 36.6 & 19.9 & 65.0 & 68.5 & 48.8 & 36.2 & 31.9 & 28.4 & 22.5 & 7.1 \\
    & CAILA$^\ast$~\citep{li2024context} & $_{\text{WACV'24}}$
    &  51.4 & 53.3 & 39.7 & 23.2 & 66.8 & 72.5 & \textbf{56.0} & 42.5 &  44.3 & 35.8 & 31.4 & 13.9  \\
    & Troika~\citep{huang2023troika} & $_{\text{CVPR'24}}$
    & 49.0 & 53.0 & 39.3 & 22.1 & 66.8 & 73.8 & 54.6 & 41.7 & 41.0 & 35.7 & 29.4 & 12.4 \\
    & CDS-CZSL~\citep{li2024context} & $_{\text{CVPR'24}}$
    & 50.3 & 52.9 & 39.2 & 22.4 & 63.9 & 74.8 & 52.7 & 39.5 & 38.3 & 34.2 & 28.1 & 11.1 \\
    & PLID~\citep{bao2023prompting} & $_{\text{ECCV'24}}$
    & 49.7 & 52.4 & 39.0 & 22.1 & 67.3 & 68.8 & 52.4 & 38.7 & 38.8 & 33.0 & 27.9 & 11.0 \\ 
   \rowcolor[rgb]{ .906,  .902,  .902}\cellcolor{white} & 
    \textbf{PLO-VLM (Ours)} &  & 49.7 & 52.8  & 39.4 & 22.3 & 67.8  & \textbf{75.6} & 53.1  & 42.0  & 43.9 & \textbf{38.2}  & \textbf{32.2} & \textbf{14.5}  \\
    \rowcolor[rgb]{ .906,  .902,  .902}\cellcolor{white} & 
    \textbf{PLO-LLM (Ours)} & & 49.6 & 53.2 & 39.0 & 21.9  & \textbf{68.3} & 73.0  & 54.8 & 41.6 & \textbf{44.3}  & \textbf{37.9} & 31.2  & \textbf{14.3} \\
    \rowcolor[rgb]{ .906,  .902,  .902}\cellcolor{white} & 
    \textbf{PLO-VLM$^\dagger$ (Ours)} & & \textbf{51.6} & \textbf{53.7} & \textbf{40.2} & \textbf{23.4} & \textbf{70.3} & \textbf{75.8} & 55.3 & \textbf{43.6} &  \textbf{44.7} & \textbf{38.1} & \textbf{33.0} & \textbf{14.9}\\
    \hline
    \multirow{13}[0]{*}{\begin{sideways}Open-World\end{sideways}} & CLIP~\citep{radford2021learning} & $_{\text{ICML'21}}$ 
    & 30.1  & 14.3  & 12.8  & 3.0  & 15.7  & 20.6  & 11.2  & 2.2   & 7.5   & 4.6  & 4.0   & 0.3  \\
    & CoOp~\citep{zhou2022learning} & $_{\text{IJCV'22}}$
    & 34.6  & 9.3  & 12.3  & 2.8  & 52.1  & 31.5  & 28.9  & 13.2  & 21.0  & 4.6  & 5.5  & 0.7  \\
    & Co-CGE~\citep{mancini2022learning}& $_{\text{TPAMI'22}}$ 
    & 38.1 & 20.0 & 17.7 & 5.6 & 59.9 & 56.2 & 45.3 & 28.4 & 33.2 & 3.9 & 5.3 & 0.9 \\
    & ProDA~\citep{lu2022prompt} & $_{\text{CVPR'22}}$
    & 37.5 & 18.3 & 17.3 & 5.1 & 63.9 & 34.6 & 34.3 & 18.4 & - & - & - & - \\
    & CSP~\citep{nayak2022learning} & $_{\text{ICLR'23}}$  
    & 46.3  & 15.7  & 17.4  & 5.7  & 64.1  & 44.1  & 38.9  & 22.7  & 28.7  & 5.2  & 6.9  & 1.2  \\
    & DFSP(i2t)~\citep{lu2023decomposed} & $_{\text{CVPR'23}}$ 
    & 47.2  & 18.2  & 19.1  & 6.7  & 64.3  & 53.8  & 41.2  & 26.4  & 35.6  & 6.5  & 9.0  & 2.0  \\
    & DFSP(BiF)~\citep{lu2023decomposed}&  $_{\text{CVPR'23}}$
    & 47.1  & 18.1  & 19.2  & 6.7  & 63.5  & 57.2  & 42.7  & 27.6  & 36.4  & 7.6  & 10.6  & 2.4  \\
    & DFSP(t2i)~\citep{lu2023decomposed} & $_{\text{CVPR'23}}$ 
    & 47.5  & 18.5  & 19.3  & 6.8  & 66.8  & 60.0  & 44.0  & 30.3  & 38.3  & 7.2  & 10.4  & 2.4  \\
    & GIPCOL~\citep{xu2023gipcol} & $_{\text{WACV'24}}$
    & 48.5 & 16.0 & 17.9 & 6.3 & 65.0 & 45.0 & 40.1 & 23.5 & 31.6 & 5.5 & 7.3 & 1.3 \\ 
    & CAILA$^\ast$~\citep{li2024context} & $_{\text{WACV'24}}$
    &  \textbf{51.4} &  \textbf{20.1} & 20.9 & \textbf{8.0} & 65.1 & 59.6 & 44.8 & 29.9 &
    43.8 & \textbf{11.4} & 7.9  & 3.3\\
    & Troika~\citep{huang2023troika} & $_{\text{CVPR'24}}$
    & 48.8 & 18.7 & 20.1 & 7.2 & 66.4 & 61.2 & 47.8 & 33.0 & 40.8 & 7.9 & 10.9 & 2.7 \\
    & \textcolor{gray}{CDS-CZSL$^\star$~\citep{li2024context}} & $\textcolor{gray}{_{\text{CVPR'24}}}$
    & \textcolor{gray}{49.4} & \textcolor{gray}{21.8} & \textcolor{gray}{22.1} & \textcolor{gray}{8.5} & \textcolor{gray}{64.7} & \textcolor{gray}{61.3} & \textcolor{gray}{48.2} & \textcolor{gray}{32.3} & \textcolor{gray}{37.6} & \textcolor{gray}{8.2} & \textcolor{gray}{11.6} & \textcolor{gray}{2.7} \\
    & PLID~\citep{bao2023prompting} & $_{\text{ECCV'24}}$
    & 49.1 & 18.7 & 20.0 & 7.3 & 67.6 & 55.5 & 46.6 & 30.8 & 39.1 & 7.5 & 10.6 & 2.5 \\ 
    \rowcolor[rgb]{ .906,  .902,  .902}\cellcolor{white} &
    \textbf{PLO-VLM (Ours)}  & & 49.7 & 19.4 & \textbf{21.4} & 7.8 & \textbf{68.0} & \textbf{63.5} & \textbf{47.8} & \textbf{33.1} & \textbf{43.9} & 10.4 & \textbf{13.9} & \textbf{3.9} \\
    \rowcolor[rgb]{ .906,  .902,  .902}\cellcolor{white} &
    \textbf{PLO-LLM (Ours)} & & 49.8 & 19.1 & 20.6 & 7.5 & \textbf{68.0} & 58.2 & \textbf{48.8} & \textbf{33.8} & 43.6 & 10.4 & \textbf{13.6} & \textbf{3.9} \\
    \specialrule{0.05em}{0pt}{0pt}
    \hline
    \end{tabular}%
    }
  \caption{Performance comparison (\%) on three datasets. $\dagger$ denotes uses MoA adapter~\cite{zheng2024caila}. $\star$ indicates the use of an additional filter trick \emph{w/o} releasing codes. $\ast$ denotes reproduction results using the official released code.}  
  \label{tab:sota}%
  \vspace{-1.5em}
\end{table*}%

\vspace{-2.0em}
\subsection{Training Objectives and Inference}

\noindent\textbf{PLO-VLM.}
During training, we employ four main losses to optimize the model: multi-label loss $\mathcal{L}_{obs}$ for the pre-observing classifier and cross-entropy losses $\mathcal{L}_{s}$, $\mathcal{L}_{o}$, and $\mathcal{L}_{c}$ for state, object, and composition, respectively. The combined loss is formulated as:
\begin{equation}
\setlength\abovedisplayskip{5pt}
    \mathcal{L}_{PLO}^{VLM} = \mathcal{L}_{obs} + \mathcal{L}_{s} + \mathcal{L}_{o} + \mathcal{L}_{c}.
\setlength\belowdisplayskip{5pt}
\end{equation}
The multi-label loss optimizes the pre-observing classifier, responsible for predicting whether to initially observe the state or object primitive in the input image. It is calculated using binary cross-entropy loss for each target label:
\begin{equation}
% \begin{aligned}
\setlength\abovedisplayskip{5pt}
    \mathcal{L}_{obs} = -y_{obs} \log(\sigma(\mathcal{F}^{pre}_{obs}))+ (1-y_{obs})\log(1-\sigma(\mathcal{F}^{pre}_{obs})),
% \end{aligned}
\setlength\belowdisplayskip{5pt}
\end{equation}
where $\sigma(\cdot)$ is sigmoid function. ${y}_{obs}$ = $[y_s, y_o]$ is ground-truth multi-label target for the pre-observing classifier, where $y_s$ and $y_o$ are binary values of ground-truth state and object categories. Both $\mathcal{F}^{pre}_{obs}$ and ${y}_{obs}$ with the shape of [|$S$| + |
$O$|], where |$S$| and |$O$| are the total class numbers of states and objects.

The optimization of classification involves three independent cross-entropy losses: $\mathcal{L}_{s}$ for state, $\mathcal{L}_{o}$ for object, and $\mathcal{L}_{c}$ for composition. Taking the state as example, its loss $\mathcal{L}_{s}$ is defined as:
\begin{equation}
% \small
\setlength\abovedisplayskip{5pt}
\mathcal{L}_{s} = -\textstyle{\sum}_{S} \; y_slog(p(s|I)).
\setlength\belowdisplayskip{5pt}
\end{equation}
During inference, we pick the composition with the highest score of $p(c|I)$ as our predicted label.

\noindent\textbf{PLO-LLM.}
The loss includes the cross-entropy loss $\mathcal{L}_{step}$ for each observation step and cross-entropy loss $\mathcal{L}_{c}$ for composition: 
\begin{equation}
\setlength\abovedisplayskip{5pt}
    \mathcal{L}_{PLO}^{LLM} = \mathcal{L}_{step} + \mathcal{L}_{c}.
\setlength\belowdisplayskip{5pt}
\end{equation}
$\mathcal{L}_{c}$ is the same as in PLO-VLM, and $\mathcal{L}_{step}$ is calculated by: 
\vspace{-1em}
\begin{equation}
\setlength\abovedisplayskip{5pt}
\mathcal{L}_{step} = -\sum_{i=1}^{n}\sum_{j \in C}{y^{(i)}_{c}} \log(p(c_j^{(i)}|I, c_j^{(1)}, \dots, c_j^{(i-1)})),
\setlength\belowdisplayskip{5pt}
\end{equation}
where ${y}^{(i)}_{c}$ is the ground truth one-hot encoded composition label for the $i$-th observation at step $i$.

Since PLO-LLM includes both the step-by-step hard prompt observation and soft prompt during training, the inference involves predicting the composition label $\hat{y}_c$ by calculating the probabilities of the composition categories using the following equations:
\begin{equation}
\setlength\abovedisplayskip{5pt}
\begin{aligned}
&p(c|I) = p^{Soft}(c|I) + p^{Hard}(c|I), \\
&p^{Hard}(c|I) = \prod_{i=1}^{n} p(c^{(i)}|I, c^{(1)}, \dots, c^{(i-1)}),
\end{aligned}
\setlength\belowdisplayskip{5pt}
\end{equation}
where $p(c|I)$ is the probability of the composition category $c$ and $p^{Soft}(c|I)$ is the probability based on soft prompts, and $p^{Hard}(c|I)$ is based on hard prompts at each observing step. The composition category with the highest probability is chosen as the final predicted composition label $\hat{y}_c$. The training and inference procedures of PLO-VLM and PLO-LLM are shown in the Appendix.

\section{Experiments}

\subsection{Experiment Settings}
\label{sec:4.1}
\noindent\textbf{Datasets.}
We evaluated performance on three challenging benchmarks: \textbf{MIT-States}~\citep{isola2015discovering}: It comprises 53,753 natural images, with 115 states and 245 objects. In the closed-world setting, the search space includes 1,262 seen compositions and 300/400 unseen compositions for validation/testing. 2) \textbf{UT-Zappos}~\citep{naeem2021learning}: It consists of 50,025 images of shoes, with 16 states and 12 objects. In the closed-world experiments, we considered 83 seen compositions and 15/18 (validation/test) unseen compositions following the constraints defined in~\citep{purushwalkam2019task}. 3) \textbf{C-GQA}~\citep{yu2014fine}: It contains 453 states and 870 objects, comprising a total of 39,298 images. The dataset is divided into 5,592 seen compositions for training, and 1,040/923 unseen compositions for validation/testing, respectively. In the open-world settings, these datasets contain 28,175, 192, and 278,362 compositions, respectively.

\noindent\textbf{Metrics.}
We followed the established CZSL evaluation protocol~\citep{mancini2021open} and assessed all results using four metrics in both closed-world and open-world scenarios: 1) \textit{Seen} (\textbf{S}): measures the accuracy specifically for seen compositions. 2) \textit{Unseen} (\textbf{U}): evaluates the accuracy exclusively for unseen compositions. 3) \textit{Harmonic Mean} (\textbf{HM}): denotes the best harmonic mean between the seen and unseen accuracy, providing a comprehensive performance measure. 4) \textit{Area Under the Curve} (\textbf{AUC}): computes the area under the seen-unseen accuracy curve, capturing the overall performance characteristics over a wide range of operating points from $-\infty$ to $+\infty$.

\noindent\textbf{Implementation Details.} Our PLO models were trained and evaluated on one NVIDIA A100 GPU using PyTorch~\citep{NEURIPS2019_bdbca288}. The GPT-3.5-turbo, a variant of the GPT model, known for its impressive performance, was employed as the LLM. For the CLIP, we utilized OpenAI's resources, opting for the Vision Transformer with a base configuration of ViT-L/14. In PLO-VLM, we assigned the weight factors for $\mathcal{L}_{obs}$, $\mathcal{L}_{s}$, $\mathcal{L}_{o}$, and $\mathcal{L}_{c}$ to 1.0, 0.01, 0.01, and 1.0, respectively, as also presented in~\citep{lu2023decomposed}. In PLO-LLM, the default number of observation cues was set to 4. Further, fusion weights during the aggregation of probabilities for $p^{Soft}(c|I)$ and $p^{Hard}(c|I)$ were set to 0.7 and 0.3, respectively. Moreover, the weight factors assigned to $\mathcal{L}_{step}$ and $\mathcal{L}_{c}$ were both set to 1.0. Following~\citep{lu2023decomposed}, we used the Adam optimizer for 20 epochs on all datasets, with a learning rate of 2.5e-4 and a batch size of 64. In the open-world evaluation, we adhered to the post-training calibration method~\citep{nayak2022learning} to filter out compositions deemed infeasible. 

\vspace{-1em}
\subsection{Comparison with the State-of-the-Arts}
\noindent\textbf{Setting.} To ensure fair comparisons with existing work, we conducted a comparative analysis of PLO against several CLIP-based approaches that use the same backbone (ViT-L/14).
The methods including \textbf{CLIP}~\citep{radford2021learning}, \textbf{CoOp}~\citep{zhou2022learning}, \textbf{Co-CGE}~\citep{mancini2022learning}, \textbf{ProDA}~\citep{lu2022prompt}, \textbf{CSP}~\citep{nayak2022learning}, all versions of \textbf{DFSP}~\citep{lu2023decomposed}, \textbf{GIPCOL}~\citep{xu2023gipcol}, \textbf{CAILA}~\citep{zheng2024caila}, \textbf{Troika}~\citep{huang2023troika}, \textbf{CDS-CZSL}~\citep{li2024context} and \textbf{PLID}~\citep{bao2023prompting}.

\textbf{Quantitative Analysis.}
The results are shown in Table~\ref{tab:sota}.

It can be seen that PLO outperforms nearly all competitors by clear margins in both closed-world and open-world settings across three prevalent benchmarks. 
In the closed-world setting: 1) PLO-VLM surpasses the leading competitor, Troika (with the official adapter~\cite{houlsby2019parameter}), by \textbf{0.2\%}, \textbf{0.3\%}, and \textbf{2.1\%} in AUC (the core metric) on the MIT-States, UT-Zappos, and C-GQA datasets, respectively. PLO-VLM$^\dagger$ outperforms the existing SOTA methods on almost all metrics. 2) PLO-VLM and PLO-LLM achieve dominant results on C-GQA (the most challenging dataset), with HM scores of \textbf{32.2\%} and \textbf{31.2\%}, compared to Troika's 29.4\%. 
In the open-world setting: 1) PLO-VLM attains the highest AUC across two datasets, with scores of \textbf{7.8\%}, \textbf{33.1\%}, and \textbf{3.9\%}. The performance gains may be that open-world is a more difficult setting with no constraint on the test time search space~\cite{mancini2021open}. Our PLO can achieve better performance by implicitly contracting the search space by first observing the salient features. 2) PLO-LLM also exhibits strong performance, particularly excelling in the UT-Zappos dataset (\textbf{48.8}\% on HM metric). Notably, since the visual image properties (\eg, resolution, clarity, and consistency) of the unseen portion in the same dataset align with those of the seen data, PLO-VLM marginally outperforms PLO-LLM on these samples. However, for cross-domain data (\cf, Appendix), PLO-LLM significantly surpasses PLO-VLM.
These numerical results substantiate our motivation of empowering models with the \emph{observing} capabilities to understand visual compositions in a simple-to-complex manner, rather than modeling each composition separately.

\begin{table*}[!htpb] 
\hspace{1em}
\begin{subtable}{0.45\linewidth}
{
    \centering
    \renewcommand\arraystretch{1.0}
        \setlength\tabcolsep{11pt}
        
  {
    \begin{tabular}{|cc||cccc|}
     \thickhline
    \multicolumn{2}{|c||}{Component} & \multicolumn{4}{c|}{MIT-States} \\

    \multicolumn{1}{|c}{PEFT} & \multicolumn{1}{c||}{DO} & \multicolumn{1}{c}{S} & \multicolumn{1}{c}{U} & \multicolumn{1}{c}{HM} & \multicolumn{1}{c|}{AUC} \\
    \hline
    \hline
    \xmark & \xmark & 47.6  & 51.2  & 36.9  & 20.4  \\
    \cmark & \xmark & 47.7  & 52.8  & 38.1  & 21.3 \\
    \xmark & \cmark & 48.5  & 52.0  & 38.3  & 21.3 \\
    \cmark & \cmark & 49.7  & 52.8  & 39.4  & 22.3  \\
    \specialrule{0.05em}{0pt}{0pt}
    \hline
    \end{tabular}%
    }}
\caption{Ablation study on each component.}
\label{tab:components}
\end{subtable}
\hspace{3em}
\begin{subtable}{0.46\linewidth}
{
    \centering
    \renewcommand\arraystretch{1.0}
    \setlength\tabcolsep{11pt}
 {
    \begin{tabular}{|c||cccc|}
     \thickhline
    \multirow{2}[4]{*}[1.3ex]{\parbox{1.5cm}{ First \\ Observation}} & \multicolumn{4}{c|}{MIT-States} \\
        & \multicolumn{1}{c}{S} & \multicolumn{1}{c}{U} & \multicolumn{1}{c}{HM} & \multicolumn{1}{c|}{AUC} \\
    \hline
    \hline
    \textit{Baseline} & 47.6  & 51.2  & 36.9  & 20.4  \\
    State & 49.5  & 52.0 & 39.0    & 21.9 \\
    Object & 48.4  & 52.8  & 38.2  & 21.5 \\
    Dynamic & 49.7  & 52.8  & 39.4  & 22.3  \\
    \specialrule{0.05em}{0pt}{0pt}
    \hline
    \end{tabular}%
    }}
\caption{Ablation study on observation order.}
\label{tab:ab_order}
\end{subtable}
\vspace{-1.0em}
\caption{Ablation study on PLO-VLM.}
\vspace{-2.0em}
\label{tab:aba_plovlm}%
\end{table*}%

\subsection{Ablation Study}
In this section, we conducted ablation studies to analyze the effectiveness of each component in PLO-VLM, the impact of the observation order in PLO-VLM, the number of observation cues in PLO-LLM, and the network architectures of the based CLIP. All experiments were conducted on the MIT-States dataset under the closed-world setting\footnote{More ablation study results are left in the \textbf{Appendix}.\label{footnote:Abla}}.

\underline{\textbf{Key Components in PLO-VLM.}} 
The contribution of Parameter Efficient Fine-Tuning (PEFT) and Dynamic Observation (DO) strategies to PLO-VLM was assessed, as summarized in Table~\ref{tab:aba_plovlm}(a). Our analysis can yield the following insights: 1) \textbf{PEFT (Fine-Tuning in $En_v$)}: Incorporating PEFT into the image encoder $En_v$ marginally improves HM metric, with an increase from 36.9\% to 38.1\%. This improvement underscores the subtle yet impactful role of fine-tuning the encoder's lightweight layers. More results of PEFT are left in the \textbf{Appendix}.
2) \textbf{DO}: Implementing DO in isolation enhances the model's ability to recognize both seen and unseen compositions, as reflected in a boost of the seen metric from 47.6\% to 48.5\% and unseen metric from 51.2\% to 52.0\%. Even when compared to the SOTA DFSP method without PEFT, our DO strategy achieves a certain improvement (21.3\% \vs~20.8\% on AUC metric). This highlights the critical impact of dynamically determining observations in understanding visual compositions. 3) \textbf{Synergistic Impact of PEFT and DO}: When PEFT and DO are synergistically applied, they collectively achieve the highest AUC of 22.3\%. This composite application underlines the effectiveness of integrating both strategies, leading to optimal model performance in recognizing compositions.

\begin{figure}[!t]
    \centering
    \includegraphics[width=1.0\linewidth]{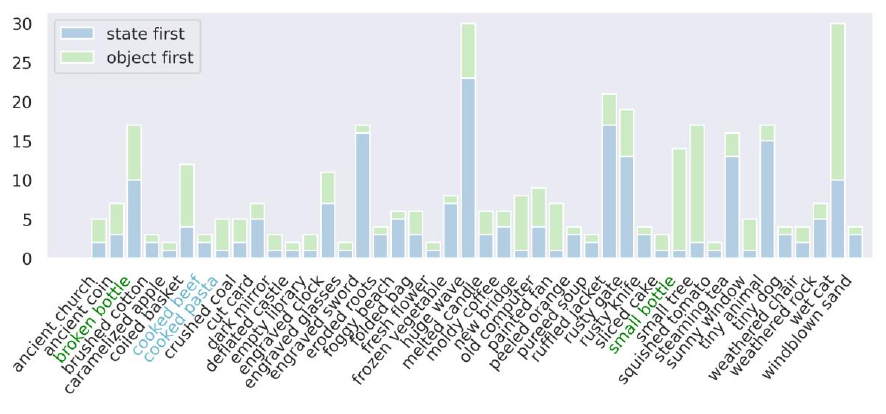}
    \vspace{-3.2em}
    \caption{Distribution of samples where state or object is first observed in PLO-VLM on MIT-States test set.}
    \label{fig:bar}
    \vspace{-0.5em}
\end{figure}

\begin{figure}
    \centering
    \includegraphics[width=\linewidth]{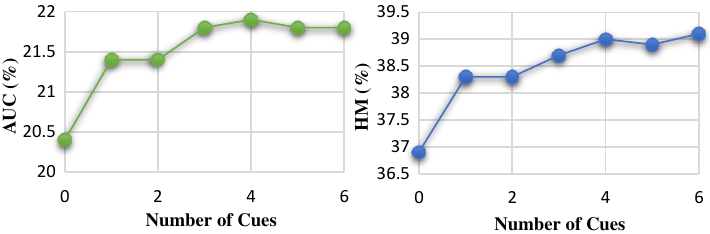}
    \vspace{-2.5em}
    \caption{Ablation study on the different number of observation cues in PLO-LLM.}
    \label{fig:llm-prompt}
    \vspace{-0.5em}
\end{figure}

\underline{\textbf{Observation Order.}} 
We explored the influence of different observation orders in PLO-VLM by using various strategies for determining the first observation: predicting the state primitive first, predicting the object primitive first, and dynamically deciding the observation order. From Table~\ref{tab:aba_plovlm}(b), we can observe: 1) The multi-step observation leads to significant performance gains against the baseline across all the metrics. 2) Dynamically deciding the observation order based on the input image yields the highest overall performance. These findings verified the effectiveness of adaptively choosing the most informative observation step based on the content of the input image, which is of importance due to the conditioned variance nature of CZSL.

Figure~\ref{fig:bar} reveals the dynamic observation strategy of PLO-VLM, which selects states or objects first based on their visual saliency in a composition. Note that in the stacked bar, blue represents the samples of ``state first'', green represents ``object first'', and the green bar DOES NOT cover blue samples. As seen, the ``\texttt{cooked pasta}'' often undergoes object-first observations owing to its distinct visuals, whereas ``\texttt{broken bottle}'' typically requires state-first recognition due to the impactful visual change presented by the state ``\texttt{broken}''. PLO-VLM's context-aware approach mirrors human cognition, leading to more precise recognition.

\underline{\textbf{Number of Observation Cues.}}
We investigated the impact of the number of observation cues in PLO-LLM. We varied the number of cues for each composition category and evaluated the model's performance accordingly. The results in Figure~\ref{fig:llm-prompt} reveal a nuanced relationship between the number of cues and the model's accuracy. As expected, increasing the number of observed cues can generally lead to improved performance, and performance is saturated when the number is 4. Thus, the number of cues is set to 4 as default.

\begin{figure}[!t]
    \centering
    \includegraphics[width=0.90\linewidth]{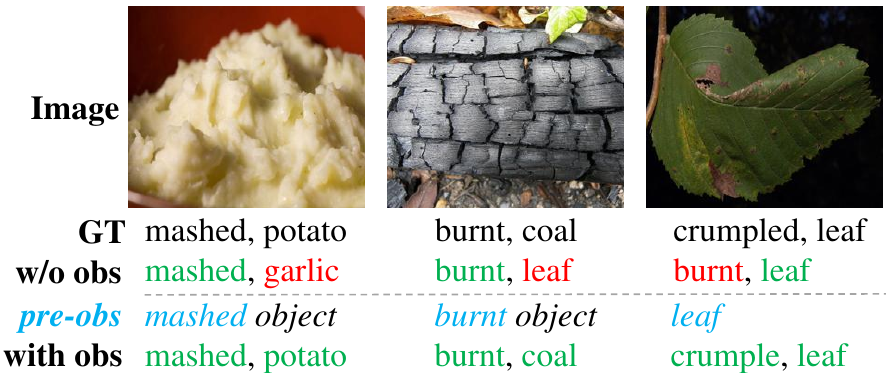}
    \vspace{-1.0em}
    \caption{Top-1 predictions of PLO-VLM with and \textit{w/o} dynamically observing under open-world setting on MIT-States. Pre-observing primitives are highlighted in \textcolor{myblue}{blue}.}
    \label{fig:obs}
    \vspace{-0.5em}
\end{figure}

\subsection{Visualization}

In this section, the validity of PLO is further illustrated by the prediction results of some samples and the cross-attention map visualization of each observation step.

\begin{figure}[!t]
\centering
\includegraphics[width=0.95\linewidth]{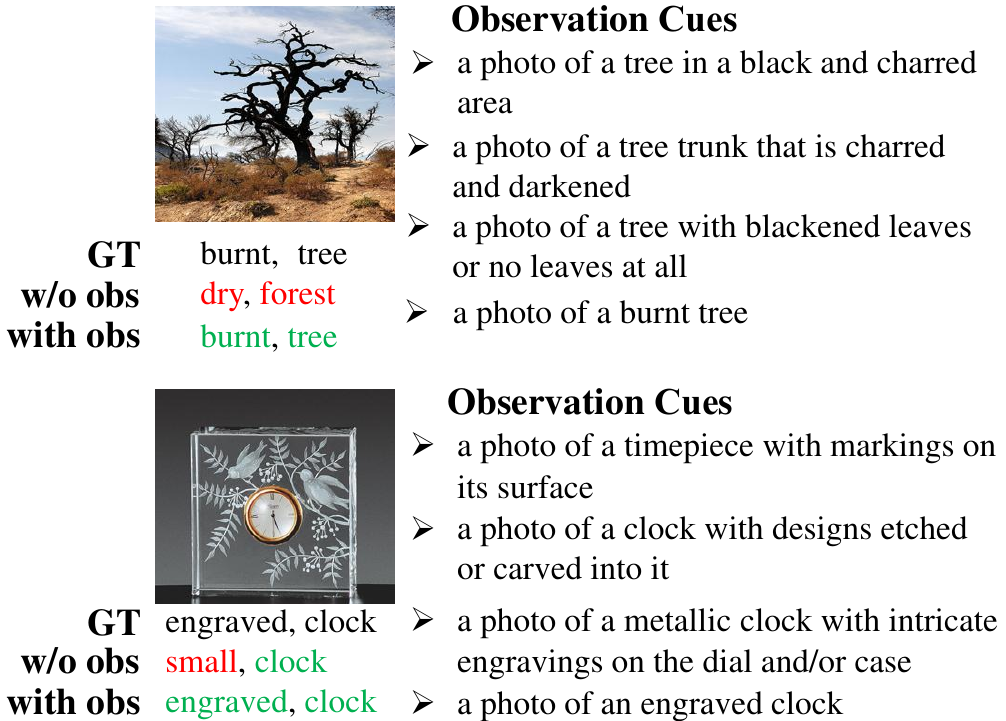}
% \vspace{-0.5em}
\caption{Top-1 predictions of PLO-LLM with and w/o multi-step observations under the closed-world setting on the MIT-States dataset. Corresponding observation cues are on the right of each image. Correct and incorrect predictions are in \textcolor{mygreen}{green} and \textcolor{red}{red}, respectively.}
\label{fig:llm_results}
% \vspace{-1em}
\end{figure}

\underline{\textbf{Top-1 Predictions.}}
We visualized top-1 predictions of PLO-VLM with and without dynamically observing in Figure~\ref{fig:obs}. In a sense, the pre-observed primitive can represent a more prominent feature in the image (\eg, ``\texttt{mashed object}'' and ``\texttt{burnt object}''). By virtue of pre-observed primitive, PLO-VLM consistently achieved accurate predictions.
In Figure~\ref{fig:llm_results}, we presented the top-1 predictions obtained by PLO-LLM, in cases where observations (obs) were utilized or not. Concurrently, we displayed the sequence of step-by-step observation cues on the right side of each image. With our meticulously designed prompt, the LLM effectively generated a progression of observation cues that transitioned from easy to hard. Through the aid of these well-organized observation cues, we empowered the model with \emph{observing} capabilities, leading to better holistic understanding and considerable performance gains.

\begin{figure}[!t]
\centering
\vspace{-0.5em}
\includegraphics[width=0.98\linewidth]{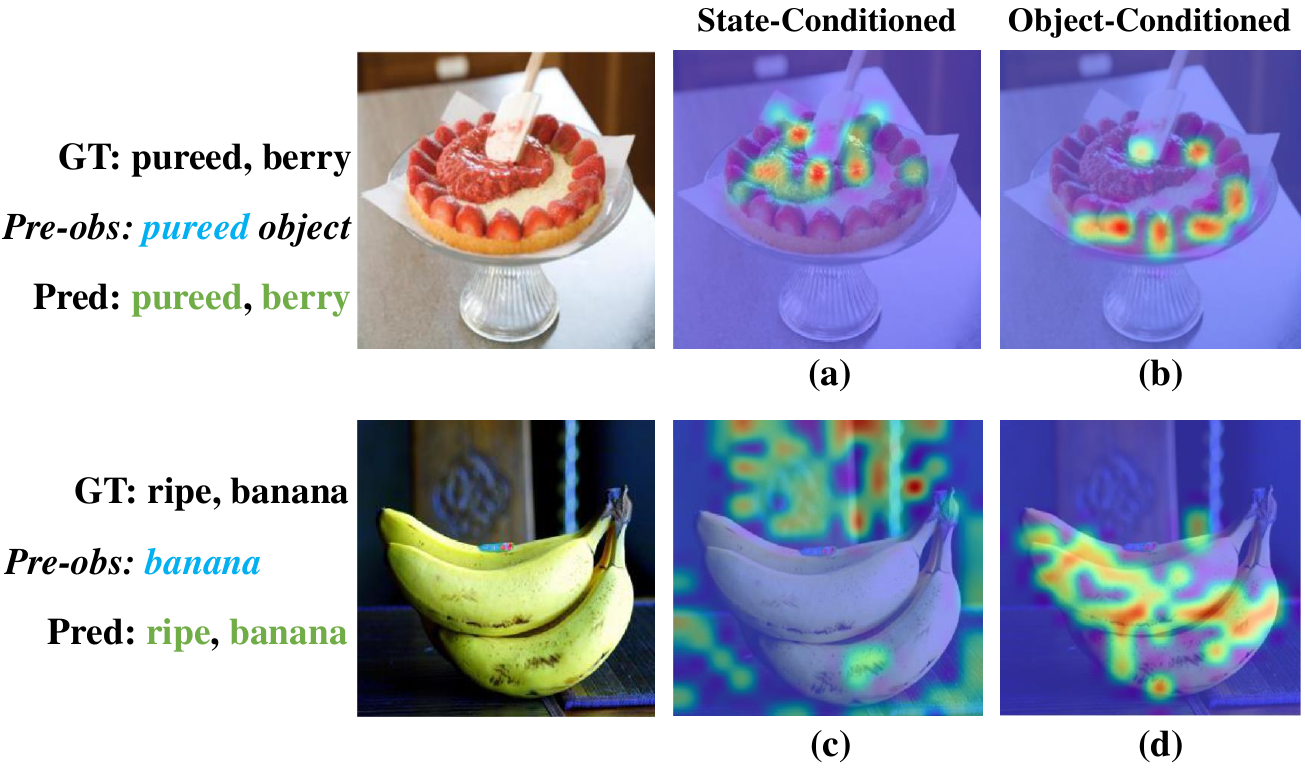}
\vspace{-1.5em}
\caption{Visualization of cross-attention maps based on different primitives in PLO-VLM.}
% \vspace{-1em}
\label{fig:vlm_ca}
\end{figure}

\begin{figure}[!t]
\centering

\includegraphics[width=1.0\linewidth]{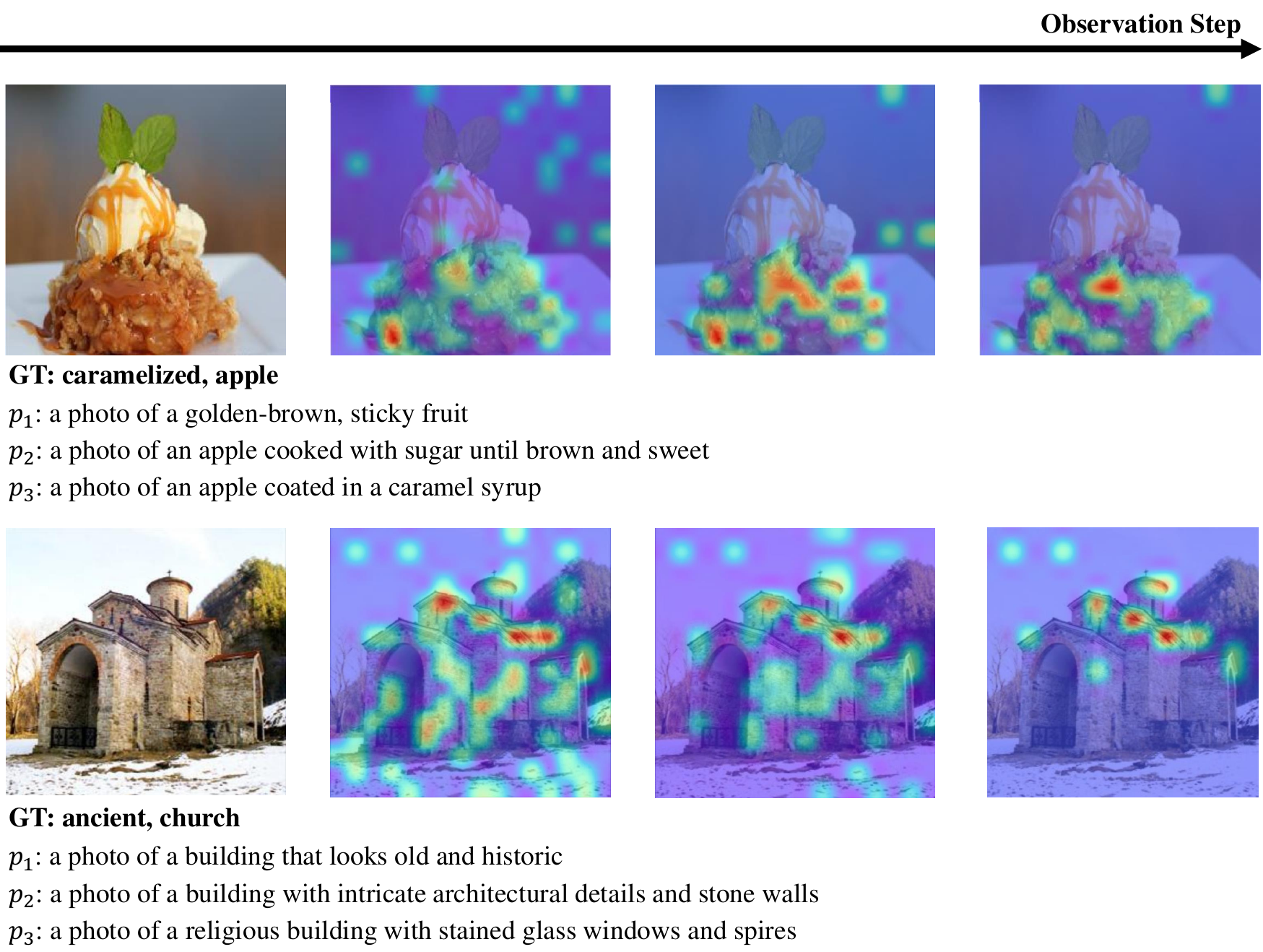}
\vspace{-1.5em}
\caption{ Visualization of cross-attention maps at different observation steps in PLO-LLM.}
% \vspace{-0.5em}
\label{fig:llm_ca}
\end{figure}

\underline{\textbf{Cross-Attention Map.}}
We displayed the cross-attention maps in Eq.~\eqref{eq:CA} of different primitives in PLO-VLM (\cf, Figure~\ref{fig:vlm_ca}) and of different steps in PLO-LLM (\cf, Figure~\ref{fig:llm_ca}) to explain our reasons for the performance improvement. 

\noindent\textbf{PLO-VLM.} 1) State First: When our model detects that the state features are more prominent, \eg, the mushy texture common to various ``\texttt{pureed}'' samples. PLO-VLM uses cross-attention to focus on the color and texture in the ``\texttt{pureed}'' areas (\cf, Figure~\ref{fig:vlm_ca}(a)). This allows it to correctly identify the composition as ``\texttt{pureed berry}''. Methods that rely solely on object features might misclassify this image as ``\texttt{fresh berry}'' (\cf, Figure~\ref{fig:vlm_ca}(b)). 2) Object First: Conversely, when object features are more discernable, such as the crescent shape of a banana, the model also applies cross-attention to look for signs of ripeness, like brown spots, to determine the composition as ``\texttt{ripe banana}'' (\cf, Figure~\ref{fig:vlm_ca}(c)). In this scenario, focusing on less prominent state features may lead to misclassification (\cf, Figure~\ref{fig:vlm_ca}(d)). By recognizing the most salient state/object automatically, our method bypasses the drawbacks of fixed order strategies, leading to more accurate recognition of compositions.

\noindent\textbf{PLO-LLM.} It can be seen that the regions of interest gradually coalesce as the steps progress. Taking the composition ``\texttt{caramelized apple}'' (top row) as an example, with the ordered text cues (\ie, ``golden-brown'',  ``with sugar'', and ``caramel syrup''), our model gradually attends to the distinctive regions of ``\texttt{caramelized apple}'', suggesting the explainability and effectiveness.

\section{Conclusion}
In this paper, we presented PLO, a novel solution for addressing the challenges of conditioned variance in CZSL. Unlike existing methods modeling each composition separately, PLO effectively captures the interactions between state and object primitives as well as graduated descriptions by automatically determining the observation cue order, leading to a holistic understanding of visual concepts for each composition. The two variants, PLO-VLM and PLO-LLM, harness the power of VLMs and LLMs to refine image representations in a simple-to-complex manner. Experimental results on three gold-standard CZSL benchmarks demonstrated their superiority over existing frameworks. PLO opens a new avenue for CZSL from the perspective of endowing models with \emph{observing} capabilities, and we hope it will pave the way for future research.

\clearpage
\section{Acknowledgments}

This work was supported by the National Key Research and Development Program of China (2024YFB3312900), Zhejiang Provincial Natural Science Foundation of China (No.~LD25F020001), and the Fundamental Research Funds for the Central Universities (226-2025-00057). Long Chen was supported by the Hong Kong SAR RGC Early Career Scheme (26208924), the National Natural Science Foundation of China Young Scholar Fund (62402408), and the HKUST Sports Science and Technology Research Grant (SSTRG24EG04).
This research was partially conducted by ACCESS – AI Chip Center for Emerging Smart Systems, supported by the InnoHK initiative of the Innovation and Technology Commission of the Hong Kong Special Administrative Region Government.

\bibliographystyle{ACM-Reference-Format}
\bibliography{sample-base}

\clearpage
\appendix

\section*{Appendix}
\appendix
This appendix is organized as follows:
\pagestyle{empty}
\thispagestyle{empty} 
\begin{itemize}
\item The training and inference procedures of PLO-VLM and PLO-LLM are detailed in Sec.~\ref{sec:train}.
\item The observation generation prompts mentioned in Sec.~\textcolor{red}{3.3} are illustrated in Sec.~\ref{sec:a}.
% \item Implementation details referenced in Sec.~\textcolor{red}{4.1} are provided in Sec.~\ref{sec:c}.
\item Additional quantitative results are presented in Sec.~\ref{sec:d}.
\item Additional qualitative results are displayed in Sec.~\ref{sec:e}.
\item Discussion between two PLO variants is shown in Sec.~\ref{sec:f}.
\item Limitations are considered in Sec.~\ref{sec:g}.
\end{itemize}

\section{The procedures of Training and Inference}
\label{sec:train}
To provide a clearer understanding of our training and inference processes, we detail the specific operational steps of PLO-VLM and PLO-LLM in Algorithm~\ref{alg:vlm} and Algorithm~\ref{alg:llm}, respectively.

\section{Observation Cue Generation}
\label{sec:a}
We present the prompt designed to generate a series of observation cues that fed into the text encoder of CLIP within the framework of PLO-LLM. The prompt's structure is as follows:
 \lstset{
  basicstyle=\ttfamily,
  columns=fullflexible, % if you want flexible inter-character spacing
  % other settings...
}

\begin{lstlisting}[numbers=none,breaklines=true]
Setting: In compositional classification, we use the hostile prompt "a photo of [state] [object]" and compute similarities between images and prompts to determine composition category: [state] [object].
Q: How can you identify a photo of the composition "mashed banana" ? Please provide step-by-step observation prompts from easy to hard, where each step builds upon the previous one. Note that the last observation prompt is "a photo of mashed banana".
A: Let's observe it step by step!
   Four observation prompts:
   - a photo of yellow, mushy substance
   - a photo of a fruit that has been mashed into a paste
   - a photo of a soft and creamy mixture made from bananas
   - a photo of mashed banana
Q: How can you identify a photo of  the composition "{STATE CLASS} {OBJECT CLASS}" ? Please provide step-by-step observation prompts from easy to hard, where each step builds upon the previous one. Note that the last observation prompt is "a photo of {STATE CLASS} {OBJECT CLASS}".
A: Let's observe it step by step!
   Four observation prompts:
\end{lstlisting}

\begin{algorithm}[!htpb]
\caption{PLO-VLM Training and Inference}
\label{alg:vlm}
\begin{algorithmic}[1] % The [1] ensures line numbers are displayed
\STATE {\bfseries Input:} Training data $DS = \{(I, c)\}$
\STATE {\bfseries Output:} Optimal parameters for PLO-VLM
\STATE {\bfseries Initialization:} Initialize weights

\STATE \texttt{// Training Process}

\WHILE{not converged}
    \STATE Sample a batch from $DS$ with images $\{I_k\}_{k=1}^{n}$ and their corresponding labels $\{c_k\}_{k=1}^{n}$
    \STATE Extract $\mathbf{v}$, $\mathbf{t}_s$, $\mathbf{t}_o$, and $\mathbf{t}_c$
    \STATE Use the pre-observing detector $\mathcal{F}^{pre}_{obs}(\mathbf{v}, \mathbf{t}_s, \mathbf{t}_o)$ to determine observation order
    \STATE Compute $\mathcal{L}_{obs}$
    \IF{observing object first}
        \STATE Obtain $\mathbf{t}^{pre}_{obs} \leftarrow \mathbf{t}_o$, $\mathbf{t}^{post}_{obs} \leftarrow \mathbf{t}_s$
        \STATE Compute refined $\widetilde{\mathbf{v}}$ via CA, based on $\mathbf{t}^{pre}_{obs}$
        \STATE Compute $p(s|I) = \pi(\mathcal{S}(\widetilde{\mathbf{v}}, \mathbf{t}^{post}_{obs})$
        \STATE Compute loss for state: $\mathcal{L}_{s}$
    \ELSE
        \STATE Obtain $\mathbf{t}^{pre}_{obs} \leftarrow \mathbf{t}_s$, $\mathbf{t}^{post}_{obs} \leftarrow \mathbf{t}_o$
        \STATE Compute the refined image representation $\widetilde{\mathbf{v}}$ via CA, based on $\mathbf{t}^{pre}_{obs}$
        \STATE Compute $p(o|I) = \pi(\mathcal{S}(\widetilde{\mathbf{v}}, \mathbf{t}^{post}_{obs})$
        \STATE Compute loss for object: $\mathcal{L}_{o}$
    \ENDIF
    \STATE Compute $p(c|I) = \pi(\mathcal{S}(\widetilde{\mathbf{v}}, \mathbf{t}_c)$ and $\mathcal{L}_{c}$
    \STATE Calculate total loss: $\mathcal{L}_{PLO}^{VLM} = \mathcal{L}_{obs} + \mathcal{L}_{s} + \mathcal{L}_{o} + \mathcal{L}_{c}$
    \STATE Update weights using $\mathcal{L}_{PLO}^{VLM}$
\ENDWHILE

\STATE \texttt{// Inference Process}

\FOR{each test image $I$}
    \STATE Compute $p(c|I)$
    \STATE Predict label: $\hat{y}_c = \text{argmax } p(c|I)$
\ENDFOR
\end{algorithmic}
\end{algorithm}

\begin{algorithm}
\caption{PLO-LLM Training and Inference}
\label{alg:llm}
\begin{algorithmic}[1] % The [1] ensures line numbers are displayed
\STATE {\bfseries Input:} Training data $DS = \{(I, c)\}$
\STATE {\bfseries Output:} Optimal parameters for PLO-LLM
\STATE {\bfseries Initialization:} Initialize weights
\STATE {\bfseries Observation Cues Generation:} 
\FOR{each composition category $c \in C$}
    \STATE Generate observation cues $\mathcal{P}_c$ using LLM
\ENDFOR

\STATE \texttt{// Training Process}

\WHILE{not converged}
    \STATE Sample a batch from $DS$ with images $\{I_k\}_{k=1}^{n}$ and their corresponding labels $\{c_k\}_{k=1}^{n}$
    \FOR{each observation step $i$}
        \STATE Extract $\mathbf{v}^{(i)}$ and $\mathbf{t}^{(i)}_{obs}$
        \STATE Compute the refined image representation $\widetilde{\mathbf{v}}^{(i)}$ via CA and update $\mathbf{v}^{(i+1)}$
        \STATE Compute probability: $p(c^{(i)}|I, c^{(1)}, \dots, c^{(i-1)})$
    \ENDFOR
    \STATE Calculate total loss: $\mathcal{L}_{PLO}^{LLM} = \mathcal{L}_{step} + \mathcal{L}_{c}$
    \STATE Update weights using $\mathcal{L}_{PLO}^{LLM}$
\ENDWHILE

\STATE \texttt{// Inference Process}

\FOR{each test image $I$}
    \STATE Compute $p^{Soft}(c|I)$
    \FOR{each observation step $i$}
        \STATE Update $p^{Hard}(c|I)$ using hard prompts
    \ENDFOR
    \STATE Predict label: $\hat{y}_c = \text{argmax } p(c|I)$
\ENDFOR
\end{algorithmic}
\end{algorithm}

The prompt is divided into four individual parts: setting, constraint, example, and question:
\begin{itemize}
\item \textbf{Setting:} The configuration setting text (\ie, ``\texttt{In compositional classification...}'') establishes a specific context and roles for the Large Language Models (LLMs) to operate within.

\item \textbf{Constraint:} The constraint (\ie, ``\texttt{Note that...}'') outlines some limitations or constraints on the output generated by the LLMs.
\item \textbf{Example:} The provided instance (\ie, the example of ``\texttt{mashed banana}'') functions as a guiding paradigm for the model to generate analogous output in the manner of in-context learning~\citep{brown2020language,liu2021makes,shao2025micas}.
\item \textbf{Question:} The question (\ie, ``\texttt{How can you identify...}'') instructs the model to devise observation cues tailored to the specific composition category under consideration. 
\end{itemize}

\begin{table}[!t]
  \centering
    \renewcommand\arraystretch{1.0}
    \setlength\tabcolsep{3pt}
  {
  \scalebox{1.0}{
    \begin{tabular}{|l||cccc|cccc|}
    \thickhline
    \multirow{2}[4]{*}[1.0ex]{\parbox{1cm}{ \centering Method}} & \multicolumn{4}{c|}{C-GQA} & \multicolumn{4}{c|}{MIT-States} \\
       
& \multicolumn{1}{c}{S} & \multicolumn{1}{c}{U} & \multicolumn{1}{c}{HM} & \multicolumn{1}{c|}{AUC}  & \multicolumn{1}{c}{S} & \multicolumn{1}{c}{U} & \multicolumn{1}{c}{HM} & \multicolumn{1}{c|}{AUC} \\
    \hline
    \hline
    DFSP~\citep{lu2023decomposed} & 49.2 & 28.4 & 26.6 & 11.7 & 45.7 & 42.8 & 33.9 & 16.8 \\
    PLO-VLM & 50.8 & 28.7 & 28.7 & 12.5 & 46.2 & 42.4 & 35.3 & 17.2 
 \\
    PLO-LLM & \textbf{52.5} & \textbf{32.1} & \textbf{32.9} & \textbf{15.1} & \textbf{50.5} & \textbf{43.6} &	\textbf{37.9} & \textbf{19.4}
 \\
    \specialrule{0.05em}{0pt}{0pt}
    \hline
    \end{tabular}%
    }}
    \caption{Cross-domain evaluation on C-GQA and MIT-States.}
    \label{tab:cross}
    \vspace{-1.5em}
\end{table}

\begin{table}[!t]
  \centering
  \setlength\tabcolsep{10pt}
   {
        \scalebox{1.0}{ \begin{tabular}{|l||cccc|}
    \thickhline
    \multirow{2}[4]{*}[1.1ex]{\parbox{1.5cm}{Strategy}} & \multicolumn{4}{c|}{MIT-States} \\
         & \multicolumn{1}{c}{S} & \multicolumn{1}{c}{U} & \multicolumn{1}{c}{HM} & \multicolumn{1}{c|}{AUC} \\
    \hline
    \hline
    CLS-DES~\cite{menon2023visual} & 47.0 & 51.8 & 37.3 & 20.3  \\
    PLO-LLM & 49.6 & 53.2 & 39.0 & 21.9  \\
    \specialrule{0.05em}{0pt}{0pt}
    \hline
    \end{tabular}%
    }}
    \caption{Comparison of LLM-based strategies on the MIT-States dataset.}
  \label{tab:llm}%
  \vspace{-2.0em}
\end{table}%

\begin{table*}[!t]
  \centering
    \setlength\tabcolsep{8pt}
   {
    \begin{tabular}{|lr||cccc|cccc|cccc|}
    \thickhline
    \multicolumn{2}{|l||}{\multirow{2}{*}{Strategy}} & \multicolumn{4}{c|}{MIT-States} & \multicolumn{4}{c|}{UT-Zappos} & \multicolumn{4}{c|}{C-GQA} \\
    &      & S     & U     & HM    & AUC   & S     & U     & HM    & AUC   & S     & U     & HM    & AUC \\
    \hline
    \hline
    % \multicolumn{12}{l}{\textit{Closed-world Setting}}\\  
    % \hline
    Adapter~\citep{houlsby2019parameter} &  & 49.7  & 52.8  & 39.4  & 22.3 & 67.8  & 75.6 & 53.1  & 42.0  & 43.9 & 38.2  & 32.2 & 14.5  \\
    LoRA~\citep{hu2022lora}  & & 50.2 & 52.9 & 38.6 & 22.3 & 69.1 & 73.8 & 55.6 & 43.2 &  42.8 & 38.3 & 31.8 & 14.2\\
    MoA~\citep{zheng2024caila}  & & 51.6 & 53.7 & 40.2 & 23.4 & 70.3 & 75.8 & 55.3 &  43.6 & 44.7 & 38.1 & 33.0 & 14.9\\
    \specialrule{0.05em}{0pt}{0pt}
    \hline
    \end{tabular}%
    }
  \caption{Ablation study on different PEFT strategies in closed-world settings.}
  \label{tab:peft_ablation}
\end{table*}%

In addition, this process adheres to the Chain-of-Thought concept~\citep{kojima2022large,zhang2022automatic,li2025relation}, generating observations step by step from easy to hard through the utilization of a special prompt ``\texttt{Let's observe it step by step!}''.

\noindent\textbf{Filter strategies}. When the query contains an unreasonable composition category, it is always difficult for LLM to generate the corresponding (or the same) format as those in-context examples given by the instruction, \eg, LLM may reply ``For this concept, which is nonsensical and not realistically possible''. We have used two filter strategies for those unreasonable composition queries:
\begin{itemize}
    \item Unreasonable observation replace. We apply strict format matching (item numbers and the prefix ``a photo of'') to filter out unreasonable categories and replace them with hard prompts. Specifically, we use the ``a photo of [state] [object]'' to replace the generated observations/descriptions of the unreasonable composition category.
    \item Unfeasible categories filter. During inference, existing state-of-the-art methods also define the list of unfeasible categories and filter out unfeasible categories during inference. In our experiments, we followed the same experimental setup as prior work~\cite{mancini2021open}.
\end{itemize}

Note that our strategy DOES NOT pose the risk of information leak. The reasons are: 1) We use exactly the same setting following previous works~\cite{menon2023visual,li2023zero}. Concretely, the state and object categories are pre-defined and remain unchanged. At testing, the \textit{same state and object categories} from training are used, but with novel compositions appearing. Since CZSL is a classification task, complete composition categories must be provided at test time. Hence, for unseen compositions in testing, we deploy LLMs to generate observation sequences (hard prompts) in the same way. 2) For hard prompt generation, we only provide the state and object category information, and DO NOT offer any image information.

% \section{Implementation Details}
% \label{sec:c} 

\section{Extra Quantitative Results}
\label{sec:d}
\subsection{Cross-Domain Evaluation}

\textbf{Setting.} DFSP(i2t)~\citep{lu2023decomposed}, PLO-VLM, and PLO-LLM on the MIT-States~\citep{isola2015discovering} dataset and tested their cross-domain performance on the C-GQA dataset~\citep{yu2014fine}. We selected categories corresponding to the states and objects in MIT-States to ensure a consistent assessment. In addition, we also conducted cross-domain tests on the MIT-States dataset in reverse.

\textbf{Results.} As shown in Table~\ref{tab:cross}, PLO-LLM exhibits superior performance on both C-GQA and MIT-States datasets, particularly in the accuracy of unseen categories (32.1\% \vs~28.7\% and 43.6\% \vs~42.4\%) and consistently outperforms in HM and AUC scores. This enhancement stems from the GPT-generated sequence of observation cues in PLO-LLM, which operates independently of the visual modality, thereby bolstering cross-domain robustness.

\subsection{Comparison with the LLM-based Models}
To prove the effectiveness of step-by-step observations, following~\cite{menon2023visual}, we added cues generated by LLM in the class label (\textbf{CLS-DES}). The results are shown in Table~\ref{tab:llm}.
It can be seen that our step-by-step fused strategy can get higher performance than cue comparison at the class-level. This phenomenon might be attributed to the necessity of one-step recognition for different cues at the class level. Unlike the one-step identification of composition, progressive observation allows for a greater focus on salient features, thereby enabling more precise predictions.

\subsection{Ablation Study on PEFT}
% \textcolor{red}{TODO}
\textbf{Setting.}  To optimize the performance of PLO-VLM for CZSL tasks, we focused on enhancing CLIP's image encoder using two distinct Parameter Efficient Fine-Tuning (PEFT) strategies: Adapter~\citep{houlsby2019parameter} and LoRA~\citep{hu2022lora}. Besides, we also conducted MoA strategy~\citep{zheng2024caila} on our PLO-VLM, which reinforces both CLIP's image encoder and text encoder in the same manner as~\citep{zheng2024caila}. These PEFT strategies are incorporated into the image encoder $En_v (\cdot)$ by inserting learnable layers into its transformer blocks. This approach allows for fine-tuning on CZSL tasks while leveraging the rich feature extraction capabilities of CLIP's pre-trained model. The results are shown in Table~\ref{tab:peft_ablation}.

\textbf{Results.} From Table~\ref{tab:peft_ablation}, we can observe that all three PEFT strategies can enhance CLIP for CZSL, with varying degrees of efficacy across different datasets. Benefiting from the more parameters that can be fine-tuned, the MoA strategy of enhancing both the image encoder and text encoder can make PLO-VLM achieve the best performance over all datasets. As for mere enhancement in image encoder, LoRA outperforms Adapter in terms of both seen and unseen metrics, while Adapter demonstrates slightly better performance in HM metric in the MIT-States dataset. On the UT-Zappos dataset, LoRA significantly exceeds Adapter, notably in seen and HM metrics. For the C-GQA dataset, the performance of both strategies is closely matched, with Adapter slightly leading in the seen and HM metrics.

\begin{figure*}[!t]
    \centering
    \includegraphics[width=1\linewidth]{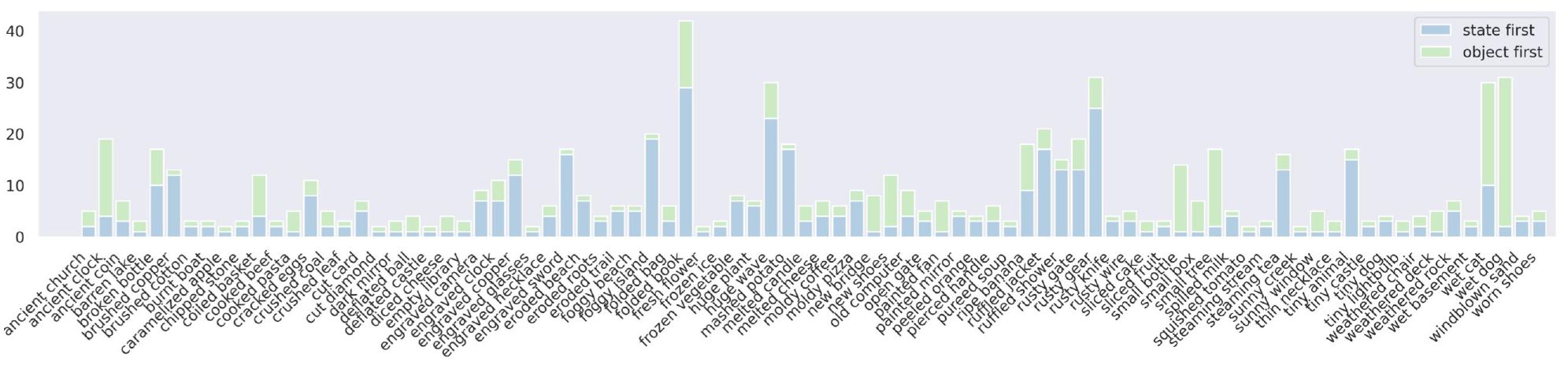}
    \vspace{-2em}
    \caption{Distribution of samples where state or object is first observed in PLO-VLM on the test set of MIT-States.}
    \label{fig:bar_supp}
\end{figure*}

\begin{table}[!t] 
  \centering
      \renewcommand\arraystretch{1.0}
    \setlength\tabcolsep{6.5pt}
     {\begin{tabular}{|l|l||cccc|}
    \thickhline
    \multirow{2}[0]{*}{Backbone} & \multirow{2}[0]{*}{Method} & \multicolumn{4}{c|}{MIT-States} \\
          &       & S     & U     & HM    & AUC \\
    \hline
    \hline
    \multirow{3}[0]{*}{ViT-B/32} & DFSP~\citep{lu2023decomposed}  & 36.7  & 43.4  & 29.4  & 13.2  \\
          & PLO-VLM & 41.1  & 44.2  & 31.3  & 14.8  \\
          & PLO-LLM & 44.2  & 47.9  & 34.3  & 17.4  \\
    \hline
    \multirow{3}[0]{*}{ViT-B/16} & DFSP~\citep{lu2023decomposed}  & 39.6  & 46.5  & 31.5  & 15.1  \\
          & PLO-VLM & 43.9  & 46.6  & 34.1  & 17.0 \\
          & PLO-LLM & 41.2  & 44.7  & 31.4  & 14.8  \\
    \hline
    \multirow{3}[0]{*}{ViT-L/14} & DFSP~\citep{lu2023decomposed}  & 46.9  & 52.0  & 37.3  & 20.6  \\
          & PLO-VLM & 49.7  & 52.8  & 39.4  & 22.3  \\
          & PLO-LLM & 49.6  & 53.2  & 39.0  & 21.9  \\
    \specialrule{0.05em}{0pt}{0pt}
    \hline
    \end{tabular}%
    }
    \caption{Ablation study on different CLIP backbones.}
  \label{tab:backbone}%
\end{table}%

\subsection{Ablation Study on Network Architectures}
% \underline{\textbf{Effect of Network Architectures.}} 
We further examined the influence of replacing CLIP backbones in our PLO-VLM and PLO-LLM. All results are reported in Table~\ref{tab:backbone}. Those consistent performance gains compared to the previous SOTA method reaffirmed the efficacy and robustness of the proposed methodology. For all other experiments, we chose ViT-L/14 by default.

\subsection{Observation Order in PLO-VLM}
The bar chart in Figure~\ref{fig:bar_supp} offers an extensive visualization of the frequency distribution for states and objects observed first within various composition categories on the MIT-States dataset. By presenting such a comprehensive statistic, we aim to provide an insightful reference that can aid future research endeavors in compositional learning and related fields.

\begin{table*}[!t]
  \centering
  \renewcommand\arraystretch{1.0}
  \setlength\tabcolsep{8pt}
  {
    \begin{tabular}{|l||cccc|cccc|cccc|}
     \thickhline
    \multirow{2}{*}{$K$-time} & \multicolumn{4}{c|}{MIT-States} & \multicolumn{4}{c|}{UT-Zappos} & \multicolumn{4}{c|}{C-GQA} \\
      & S     & U     & HM    & AUC   & S     & U     & HM    & AUC   & S     & U     & HM    & AUC \\
    \hline
    \hline
    1 & 49.7  & 52.8  & 39.4  & 22.3  & 67.8  & 75.6  & 53.1  & 42.0  & 43.9  & 38.2  & 32.2  & 14.5 \\
    2 & 49.6  & 53.0  & 39.4  & 22.3  & 68.1  & 75.3  & 53.4  & 42.1  & 43.7  & 38.1  & 32.0  & 14.4 \\
    3 & 49.8  & 52.7  & 39.3  & 22.2  & 68.2  & 75.5  & 53.0  & 42.2  & 43.8  & 38.1  & 32.1  & 14.4 \\
    4 & 49.7  & 52.7  & 39.3  & 22.2  & 68.0  & 75.5  & 53.2  & 42.1  & 43.5  & 38.2  & 32.0  & 14.3 \\
    5 & 49.9  & 52.8  & 39.4  & 22.3  & 67.8  & 75.8  & 52.9  & 42.0  & 43.7  & 38.3  & 32.1  & 14.4 \\
    \hline
    $Avg$   & 49.74  & 52.80  & 39.36  & 22.26  & 67.98  & 75.54  & 53.12  & 42.08  & 43.72  & 38.18  & 32.08  & 14.40  \\
    $Stdev$ & 0.11  & 0.12  & 0.05  & 0.05  & 0.18  & 0.18  & 0.19  & 0.08  & 0.15  & 0.08  & 0.08  & 0.07  \\
    \specialrule{0.05em}{0pt}{0pt}
    \hline
    \end{tabular}%
    }
\caption{The $K$-times experiment of PLO-VLM in closed-world setting.}
\label{tab:ktimes}%
\vspace{-1em}
\end{table*}%

\subsection{Observation Order in PLO-LLM}

\begin{table}[!t]
  \centering
    \setlength\tabcolsep{10pt}
   {
    \begin{tabular}{|l||cccc|}
    \thickhline
    \multirow{2}[4]{*}[1.3ex]{\parbox{1.5cm}{Order}} & \multicolumn{4}{c|}{MIT-States} \\
    & \multicolumn{1}{c}{S} & \multicolumn{1}{c}{U} & \multicolumn{1}{c}{HM} & \multicolumn{1}{c|}{AUC} \\
    \hline
    \hline
    (1, 2, 3)	& 49.6	& 53.2	& 39.0	& 21.9 \\
    (1, 3, 2)	& 48.8	& 52.9	& 38.7	& 21.8 \\
    (2, 1, 3)	& 48.6	& 53.1	& 38.3	& 21.6 \\
    (2, 3, 1)	& 48.8	& 53.3	& 38.4	& 21.7 \\
    (3, 1, 2)	& 49.0	& 53.1	& 38.2	& 21.6 \\
    (3, 2, 1)	& 48.8	& 52.3	& 38.5	& 21.4 \\
    \specialrule{0.05em}{0pt}{0pt}
    \hline
    \end{tabular}%
    }\caption{Ablation study on cue orders in PLO-LLM.}
\label{tab:aba_order_cue}%
\end{table}%

We also conducted ablation study on the order of cues in Table~\ref{tab:aba_order_cue}, revealing that the sequence slightly impacts performance. The last cue is fixed as ``a photo of {state} {object}'' and we change the order of the previous three cues generated by LLM. As seen, when cues follow an easy-to-difficult progression, PLO-LLM achieves better performance. Conversely, a reverse order, from difficult to easy, results in the lowest performance. This suggests that starting with simpler cues allows the model to gradually build a foundational understanding.

\subsection{Robustness of PLO} To analyze the robustness of our PLO, we provided the performance deviations of PLO-VLM by $K$-times experiments with the same parameters. The results are shown in Table~\ref{tab:ktimes}. As seen, PLO-VLM can stably outperform existing methods (\cf, Table~\ref{tab:sota}).

\subsection{Ablation Study on the Hard Prompt and Soft Prompt}

\begin{table}[!t]
  \centering
    \setlength\tabcolsep{12pt}
   {
    \begin{tabular}{|l||cccc|}
    \thickhline
     \multirow{2}[4]{*}[1.3ex]{{$\lambda$}} & \multicolumn{4}{c|}{MIT-States} \\
    & \multicolumn{1}{c}{S} & \multicolumn{1}{c}{U} & \multicolumn{1}{c}{HM} & \multicolumn{1}{c|}{AUC} \\
    \hline
    \hline
    0.0	&  40.3 & 42.9 & 30.1 & 14.0  \\
    0.3	& 47.1 & 52.9 & 35.4 & 20.7 \\
    0.5	& 48.3 & 52.9 & 37.4 & 21.4 \\
    0.7	& 49.6 & 53.2 & 39.0 & 21.9 \\
    1.0	& 49.0 & 52.9 & 38.5 & 21.6 \\
    \specialrule{0.05em}{0pt}{0pt}
    \hline
    \end{tabular}%
    }\caption{Ablation study on different fusion weights of the hard and soft prompts.}
\label{tab:fusion_w}%
\end{table}%

Hard prompts generated by LLM can provide structured guidance for refining visual features, but also face limitations such as potential misalignment with CLIP's semantic space and datasets' underlying distribution. Soft prompts offer flexibility, enabling context-aware adjustments and leveraging CLIP's broad knowledge for task-specific applications with learnable tokens~\cite{zhou2022learning}. As seen in Table~\ref{tab:fusion_w}, we studied the effect of fusion weight ($\lambda$ for hard prompts and $1-\lambda$ for soft prompts). From the results we can observe that combining hard and soft prompts together leads to the best performance. $\lambda$ is set to be 0.7 by default.

\begin{figure*}[!t]
    \centering
    \includegraphics[width=0.95\linewidth]{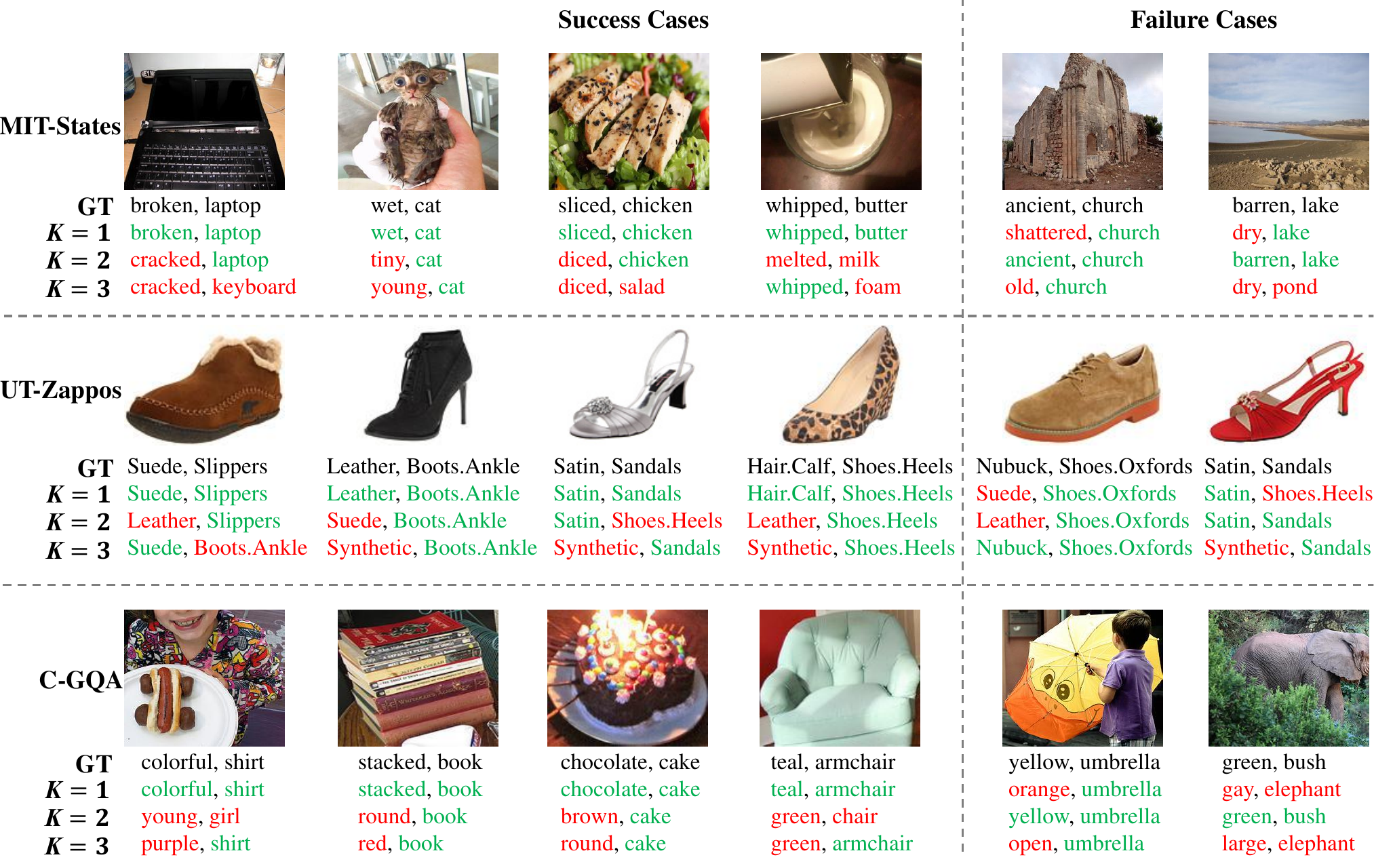}
    \caption{Top-$K$ predictions on for randomly selected cases from three datasets. The top and bottom rows show the results of closed-world settings, respectively. Correct and incorrect predictions are highlighted in \textcolor{mygreen}{green} and \textcolor{red}{red}, respectively.}
    \label{fig:cases_supp}
\end{figure*}

\begin{figure*}[!t]
    \centering
    \includegraphics[width=0.95\linewidth]{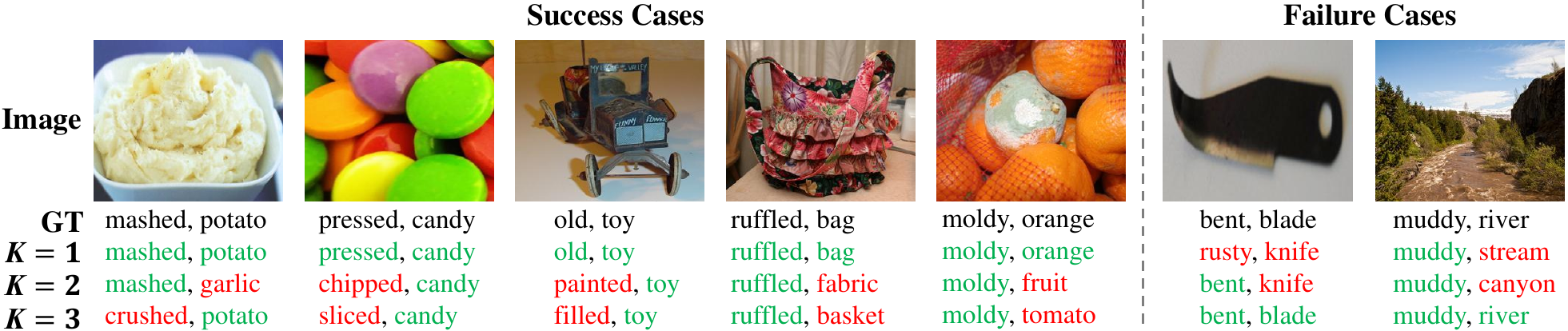}
    \caption{Top-$K$ predictions on for randomly selected cases from MIT-States. The top and bottom rows show the results of open-world settings, respectively. Correct and incorrect predictions are highlighted in \textcolor{mygreen}{green} and \textcolor{red}{red}, respectively.}
    \label{fig:cases}
\end{figure*}

\section{Extra Qualitative Results}
\label{sec:e}

\subsection{Success and Failure Cases}

Figure~\ref{fig:cases_supp} displays the top-$K$ predictions from our PLO-VLM on the MIT-States, UT-Zappos, and C-GQA datasets under closed-world setting. Successful cases, highlighted in green, demonstrate precise model predictions, exemplified by instances such as ``\texttt{sliced chicken}''. Notably, some failures, marked in red, such as ``\texttt{gray elephant}'', do not align with the ground truth (GT), yet still represent existing compositions within the image. These results underscore our model's capability to identify state-object compositions, affirming its effectiveness in interpreting complex visual scenes despite occasional misalignments with the GT. We also provided top-$K$ predictions of PLO-VLM in the open-world settings in Figure~\ref{fig:cases}. Our progressive observation strategy enables a thorough grasp of compositions, which is particularly effective in recognizing unseen compositions by bridging the gap between base and novel categories. Notably, even when our model's top-1 prediction is not exactly matched, it still accurately predicts the state primitive, and the object primitive also appears to be reasonable (\eg, \texttt{muddy} \texttt{stream}). These results underscore PLO's effectiveness in capturing pertinent visual cues and enabling comprehensive composition understanding.

\subsection{Visualization Analysis of PLO-VLM} 

\begin{figure}[!t]
\centering
\includegraphics[width=1.0\linewidth]{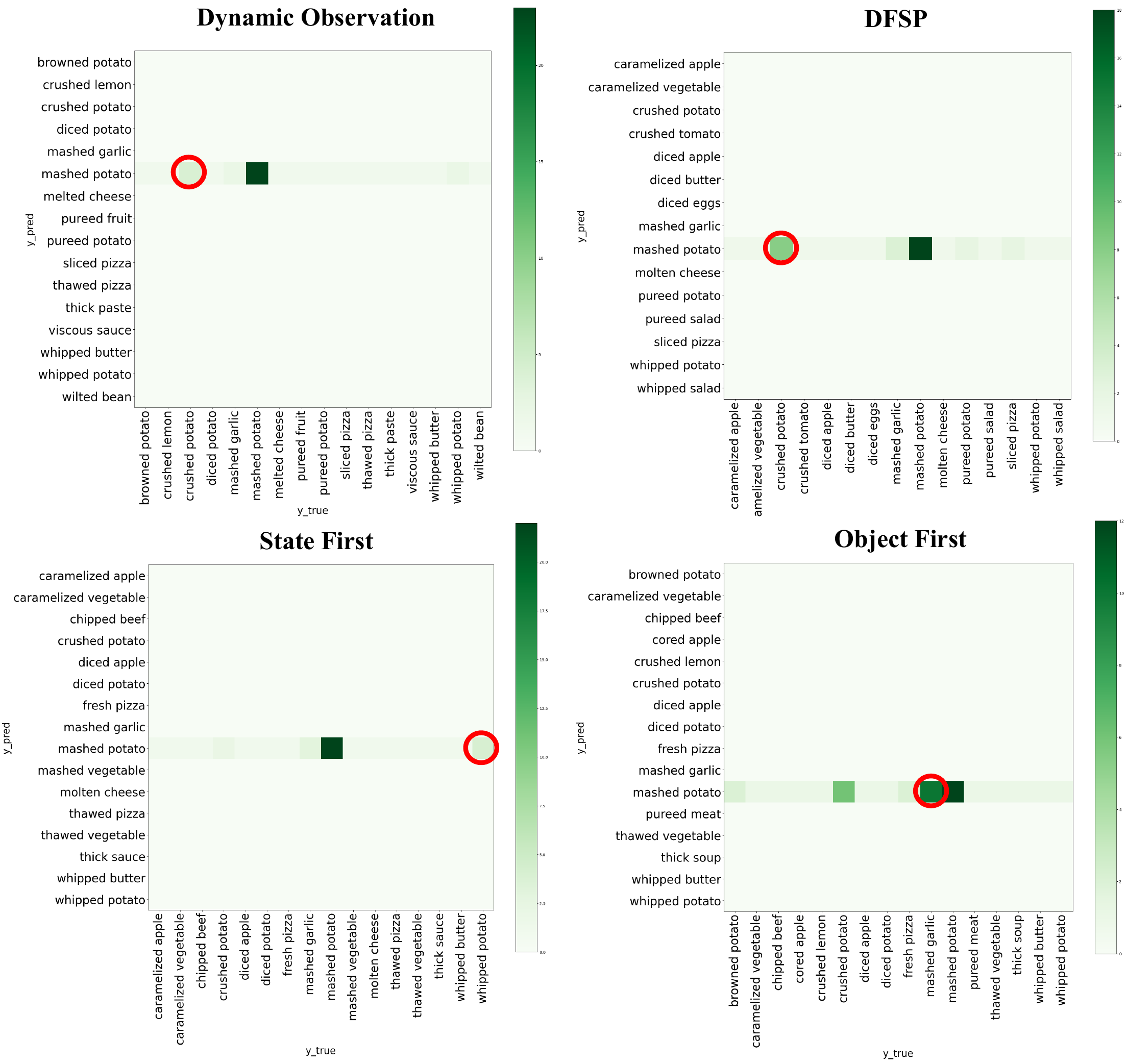}
\caption{Confusion matrix results for different models on test samples labeled as ``mashed potato''.}
\label{fig:cm}
\end{figure}

As mentioned above, we emphasize that identifying a ``mashed object'' instead of directly determining the entire composition leads to better performance. Yet, not all state-first methods ensure precise object classification. Variations in mashing degrees for potatoes and bananas present challenges. Lightly mashed samples are distinguishable by potato chunks versus banana stickiness. However, heavily mashed samples may confound even expert annotators.

To validate the effectiveness of PLO-VLM, we visualized confusion matrices of different models for testing samples labeled as ``\texttt{mashed potato}'' in Figure~\ref{fig:cm}. We outline the following key points: 1) As for the ``State First'' and ``Dynamic Observation'' cases, distinguishing between similar states such as ``\texttt{whipped}'' and ``\texttt{crushed}'' poses a significant challenge. Such similar states represent more intense (whipped) or milder (crushed) variations of the ``\texttt{mashed}'' state, leading to potential errors in state prioritization. Note that even humans find it difficult to distinguish these similar states. 2) Despite the aforementioned challenges, PLO-VLM outperforms DFSP by a clear margin, indicating PLO-VLM’s capability to recognize such nuanced differences. 3) The ``Object First'' strategy might mistakenly identify objects, \eg, ``\textit{garlic}'', due to garlic often being presented in a mashed form and sharing yellow and white colors in different states. PLO-VLM addresses this by first categorizing ``mashed object'' and then analyzing state-specific traits like texture and viscosity.

\section{Discussion of two PLO Variants}
\label{sec:f}
Although the observation steps of PLO-VLM and PLO-LLM are different, they have a \textbf{common motivation} and \textbf{common progressive observation process}.
\begin{itemize}
    \item \textbf{Common Motivation}: Both methods are designed to address primitive-conditioned variances. As detailed in Sec.~\ref{sec1}, our motivation starts with the concern that PLO-VLM's two-step observation method based on visual features may overfit specific datasets. Subsequently, we further introduce PLO-LLM to enhance robustness through a flexible, multi-step observation process.
    \item \textbf{Common Progressive Observation Process}: Both models employ a sequential refinement of visual features, albeit through different strategies (two-step PLO-VLM \vs multi-step in PLO-LLM). This process utilizes the cross-attention mechanism to leverage cues from preceding steps for progressive observation, as demonstrated in Eq.~(\ref{eq:CA}) and Eq.~(\ref{ca_2}).
\end{itemize}

\section{Limitations}
\label{sec:g}
Despite the promising results demonstrated by PLO-VLM and PLO-LLM in compositional zero-shot learning, we acknowledge two principal limitations:
\textbf{1) Scope of Zero-Shot Learning}: Our approach primarily addresses the recognition of novel compositions involving already seen states and objects. It does not extend to the zero-shot recognition of entirely novel state and object categories. This limitation marks a boundary in our model's applicability, underscoring the need for future developments that can generalize to entirely unseen states and objects.
\textbf{2) Dependence on External Language Model APIs}: The efficacy of PLO-LLM is partly reliant on external language model APIs, such as those from OpenAI. This reliance introduces practical constraints, especially concerning the costs associated with API usage, which can escalate with the increase in the number of unique composition categories.

\end{document}